\documentclass[journal]{IEEEtran}
\usepackage{ifpdf}
\usepackage{cite}
\usepackage{amsmath}
\usepackage{algorithmic}
\usepackage{array}
\usepackage{fixltx2e}
\usepackage{url}
\usepackage{graphicx}     
\usepackage{subfig}       
\usepackage{float}
\usepackage{epstopdf}   
\usepackage{color}
\usepackage{multirow}

\begin{document}
%
\title{Constrained Maximum Correntropy\\ Adaptive Filtering}
%
%
%

\author{Siyuan~Peng,~\IEEEmembership{Student~Member,~IEEE,}
        Badong~Chen,~\IEEEmembership{Senior~Member,~IEEE,}
        Lei~Sun,~\IEEEmembership{Senior~Member,~IEEE,}
        Wee~Ser,~\IEEEmembership{Senior~Member,~IEEE,}
        and~Zhiping~Lin,~\IEEEmembership{Senior~Member,~IEEE}
\thanks{S. Peng, Z. Lin and W. Ser are with the School of Electrical and Electronic Engineering, Nanyang Technological University, 639798 Singapore e-mail: (PENG0074@e.ntu.edu.sg, EZPLin@ntu.edu.sg, ewser@ntu.edu.sg).}
\thanks{B. Chen is with the Institute of Artificial Intelligence and Robotics, Xi'an Jiaotong University, Xi'an 710049 China e-mail: (Corresponding author, chenbd@mail.xjtu.edu.cn).}
\thanks{ L. Sun is with the School of Information and Electronics, Beijing Institute of Technology, Beijing 100081, China e-mail: (bitsunlei@126.com).}
\thanks{Manuscript received Dec. 6, 2016.}}

%
%

\markboth{Journal of \LaTeX\ Class Files,~Vol.~, No.~, December~2016}%
{Shell \MakeLowercase{\textit{et al.}}: Bare Demo of IEEEtran.cls for IEEE Journals}
%

\maketitle

\begin{abstract}
Constrained adaptive filtering algorithms including constrained least mean square (CLMS), constrained affine projection (CAP) and constrained recursive least squares (CRLS) have been extensively studied in many applications. Most existing constrained adaptive filtering algorithms are developed under the mean square error (MSE) criterion, which is an ideal optimality criterion under Gaussian noises. This assumption however fails to model the behavior of non-Gaussian noises found in practice. Motivated by the robustness and simplicity of maximum correntropy criterion (MCC) for non-Gaussian impulsive noises, this paper proposes a new adaptive filtering algorithm called constrained maximum correntropy criterion (CMCC). Specifically, CMCC incorporates a linear constraint into a MCC filter to solve a constrained optimization problem explicitly. The proposed adaptive filtering algorithm is easy to implement and has low computational complexity, and in terms of convergence accuracy (say lower mean square deviation) and stability, it can significantly outperform those MSE based constrained adaptive algorithms in presence of heavy-tailed impulsive noises. Additionally, the mean square convergence behaviors are studied under energy conservation relation, and a sufficient condition to ensure the mean square convergence and the steady-state mean square deviation (MSD) of the proposed algorithm are obtained. Simulation results confirm the theoretical predictions under both Gaussian and non-Gaussian noises, and demonstrate the excellent performance of the novel algorithm by comparing it with other conventional methods.
\end{abstract}


%
\IEEEpeerreviewmaketitle

%
%
\section{Introduction}
%
%
%
%
\IEEEPARstart{C}{ONSTRAINED} adaptive filtering algorithms have been successfully applied in domains of signal processing and communications, such as system identification, blind interference suppression, array signal processing, and spectral analysis \cite{1,2,3,4}. The main advantage of constrained adaptive filters is that they have an error-correcting feature that can prevent the accumulation of errors (e.g. the quantization errors in a digital implementation). As a well-known linearly-constrained adaptive filtering algorithm, the \emph{constrained least mean square} (CLMS) \cite{5} is a simple stochastic-gradient based adaptive algorithm, originally conceived as an adaptive solution to a linearly-constrained minimum-variance (LCMV) filtering problem in antenna array processing \cite{6}. Other stochastic-gradient based linearly-constrained adaptive algorithms were also developed \cite{7,8,9,10,11}. Although the LMS-type algorithms are simple and computationally efficient, they may suffer from low convergence speed especially when the input signal is correlated. In order to improve the convergence rate, the \emph{constrained recursive least squares} (CRLS) algorithm was derived in \cite{12}, at the cost of higher computational complexity. Some improvements of the CRLS can be found in \cite{13,14}. Several \emph{constrained affine projection} (CAP) algorithms were also developed \cite{15,16}.

Most of the existing constrained adaptive filtering algorithms have been developed based on the common mean square error (MSE) criterion due to its attractive features, such as mathematical tractability, computational simplicity and optimality under Gaussian assumption \cite{17,18}.
However, Gaussian assumption does not always hold in real-world environments, even though it is justified for many natural noises. When the signals are disturbed by non-Gaussian noises, the MSE based algorithms may perform poorly or encounter the instability problem \cite{19,23}. From a statistical viewpoint, the MSE is insufficient to capture all possible information in non-Gaussian signals. In practical situations, non-Gaussian noises are frequently encountered. For example, some sources of non-Gaussian impulsive noises are ill synchronization in digital recording, motor ignition noise in internal combustion engines, scratches on recording disks, and lighting spikes in natural phenomena \cite{20, 21, 22}.

To deal with the non-Gaussian noise problem (which usually causes large outliers), various alternative optimization criteria have been proposed to replace the MSE criterion for developing robust adaptive filtering algorithms in the literature.
In recent years, maximum correntropy criterion (MCC) has been successfully applied in diverse domains due to its simplicity and robustness \cite{19, 23, 24, 25, 26, 27, 28, 29}. As a nonlinear and local similarity measure directly related to the probability of how similar two random variables are in the bisector neighborhood of the joint space controlled by the kernel bandwidth, correntropy is insensitive to large outliers, and is frequently used as a powerful method to handle non-Gaussian impulsive noises in various applications of engineering. For instance, Singh et al. \cite{30}  and Zhao et al. \cite{25} utilized the correntropy as a cost function to develop robust adaptive filtering algorithm for signal processing, and Chen et al. extended the original correntropy by using the generalized Gaussian density (GGD) function as the kernel, and proposed a generalized correntropy for robust adaptive filtering \cite{31}. He et al. presented a MCC-based rotationally invariant principal component analysis (PCA) algorithm for image processing \cite{32}, and also incorporated the correntropy induced metric (CIM) into MCC to develop an effective sparse representation algorithm for robust face recognition \cite{33}. Bessa et al. adopted MCC to train neural networks for wind prediction in power system \cite{34}. Hasanbelliu et al. utilized information theoretic measures (entropy and correntropy) to develop two algorithms that can deal with both rigid and non-rigid point set registration with different computational complexities and accuracies \cite{35}. However, constrained adaptive filtering based on MCC has not been studied yet in the literature. In this work, a constrained maximum correntropy criterion (CMCC) adaptive filtering algorithm is proposed for signal processing especially in presence of heavy-tailed impulsive noises.

Our main contributions in this paper are summarized as follows:
\begin{itemize}
  \item First, we develop the CMCC adaptive filtering algorithm by incorporating a linear constraint into the MCC, instead of the traditional MSE criterion, to solve a constrained optimization problem explicitly. The computational complexity analysis is also presented.
  \item Second, based on the energy conservation relation \cite{36, 37,38,39}, we analyze the mean square convergence behaviors of the proposed algorithm, and present particularly a sufficient condition to guarantee the mean square convergence and the steady-state mean square deviation (MSD) in the cases of Gaussian and non-Gaussian noises.
  \item Finally, we confirm the validity of theoretical expectations experimentally, and illustrate the desirable performance (e.g. lower MSD) of CMCC by comparing it with other methods in linear-phase system identification and beamforming application.
\end{itemize}

The rest of the paper is organized as follows. In Section II, after briefly reviewing the MCC, we develop the CMCC algorithm and analyze the computational complexity. In Section III, we study the mean square convergence of the proposed algorithm. Simulation results are then presented in Section IV. Finally, Section V gives the conclusion and discusses some work in the future. Some derivations are relegated to the Appendix.

%
%
\section{CMCC Algorithm}

\subsection{Maximum Correntropy Criterion}
\begin{figure}[!t]
\centering
\includegraphics[width=2.5in]{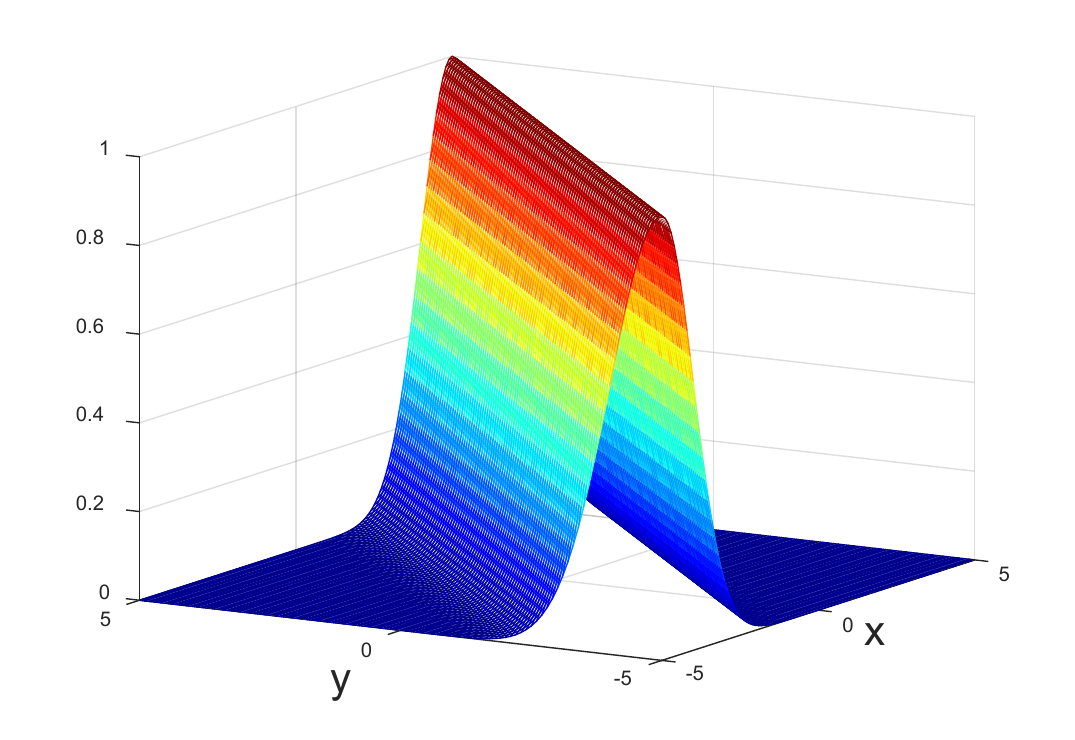}
\caption{MCC cost function in the joint space ($\sigma=1.0$).}
\label{joininmcc}
\end{figure}

As a similarity measure between two random variables $X$ and $Y$, correntropy is defined by \cite{23, 27,29,30,31}
\begin{align}\label{c_1_1}
V(X,Y)=E[\kappa(X,Y)]=\int\kappa(x,y)dF_{XY}(x,y)
\end{align}
where $E[\cdot]$ denotes the \emph{expectation operator}, $\kappa(\cdot,\cdot)$ is a \emph{shift-invariant Mercer kernel}, and $F_{XY}(x,y)$ stands for the joint distribution function of $(X,Y)$. It takes the advantage of a kernel trick that nonlinearly maps the input space to a higher dimensional feature space. In the present work, without mentioning otherwise, the kernel function of correntropy $\kappa(\cdot,\cdot)$ is the Gaussian kernel, given by
\begin{align}\label{c_1_2}
\kappa_{\sigma}(x-y)=\frac{1}{\sqrt{2\pi}\sigma}\exp\left(-\frac{(x-y)^{2}}{2\sigma^{2}}\right)
\end{align}
where $\sigma>0$ is the kernel bandwidth parameter. In most practical situations, the join distribution $F_{XY}(x,y)$ is usually unknown, and only a finite number of data samples $\{(x(n),y(n))\}^{N}_{n=1}$ are available. In these cases, the correntropy can be estimated by
\begin{align}\label{c_1_3}
\hat{V}_{N,\sigma}=\frac{1}{N}\sum\limits_{n=1}^{N}\kappa_{\sigma}(x(n)-y(n))
\end{align}
Under the \emph{maximum correntropy criterion} (MCC), an adaptive filter will be trained by maximizing the correntropy between the desired response and filter output, formulated by
\begin{align}\label{c_1_4}
\max\limits_{W}J_{MCC}=\frac{1}{N}\sum\limits_{n=1}^{N}\kappa_{\sigma}(e(n))
\end{align}
where $e(n)$ denotes the error between the desired response and filter output, and $W$ stands for the filter weight vector. Fig. \ref{joininmcc} shows the MCC cost function $\kappa_{\sigma}(x-y)$ in the joint space of $x$ and $y$. As one can see clearly, the MCC is a local similarity measure, whose value is mainly decided by the kernel function along the line $x=y$. Furthermore, from a view of geometric meaning, we can divide the space in three regions, namely Euclidean region, transition region and rectification region. The MCC behaves like 2-norm distance in the Euclidean region, similarly like a 1-norm distance in the transition region and eventually approaches a zero-norm in the rectification region, which also interprets the robustness of correntropy for outliers \cite{23,29}.
\subsection{CMCC Algorithm}
Consider a linear unknown system, with an $M$-dimensional weight vector $W^{*}=[w^{*}_{1}(n),w^{*}_{2}(n),\cdots,w^{*}_{M}(n)]^{T}$ that needs to be estimated. The measured output $d(n)$ of the unknown system at instant $n$ is assumed to be
\begin{align}\label{c_2_1}
d(n)=y^{*}(n)+\upsilon(n)=W^{*T}X(n)+\upsilon(n)
\end{align}
where $y^{*}(n)=W^{*T}X(n)$ denotes the actual output of the unknown system, $X(n)=[x_{1}(n),x_{2}(n),\cdots,x_{M}(n)]^{T}$ is the input vector, with $[\cdot]^{T}$ being the transpose operator, and $\upsilon(n)$ stands for an interference or measurement noise. Suppose the estimator is another  $M$-dimensional linear filter, with an adaptive weight vector $W(n)$. Then the instantaneous prediction error at instant $n$ is
\begin{align}\label{c_2_2}
e(n)=d(n)-y(n)=d(n)-W^{T}(n-1)X(n)
\end{align}
where $y(n)=W^{T}(n-1)X(n)$ denotes the output of the adaptive filter. For a constrained adaptive filter, a linear constraint will be imposed upon the filter weight vector as
\begin{align}\label{c_2_3}
C^{T}W=f
\end{align}
where $C$ is an $M\times K$ constraint matrix, and $f$ is a vector containing $K$ constraint values. The CLMS algorithm is derived by solving the following optimization problem \cite{5, 37}:
\begin{align}\label{c_2_4}
&\min \limits_{W}E\left[\left(d(n)-W^{T}(n-1)X(n)\right)^{2}\right]\nonumber\\
&\emph{subject to}\;\;\;\;\;C^{T}W=f
\end{align}
leading to the following weight update equation:
\begin{align}\label{c_2_5}
W(n)=&P[W(n-1)+\eta(d(n)- \nonumber \\
     &W^{T}(n-1)X(n))X(n)]+Q
\end{align}
where $\eta>0$ is the step-size parameter, $P=\textbf{I}_{M}-C(C^{T}C)^{-1}C^{T}$ with $\textbf{I}_{M}$ being an $M\times M$ identity matrix, and $Q=C(C^{T}C)^{-1}f$. \par
In this work, we use the MCC instead of MSE to develop a constrained adaptive filtering algorithm. Similar to (8), we propose the following CMCC optimization problem 
\begin{align}\label{c_2_6}
&\max\limits_{W}E\left[\kappa_{\sigma}(d(n)-W^{T}(n-1)X(n))\right]  \nonumber\\
&\emph{subject to}\;\;\;\;\;\;\;C^{T}W=f
\end{align}
and accordingly the CMCC cost $J_{CMCC}$ is 
\begin{align}\label{c_2_6_2}
J_{CMCC}=&E[\kappa_{\sigma}(d(n)-W^{T}(n-1)X(n))]+  \nonumber \\
         &\xi^{T}(n)\left(C^{T}W(n-1)-f\right)
\end{align}
%
where $\xi(n)$ is a $K\times1$ Lagrange multiplier vector. A stochastic-gradient based algorithm can thus be derived as (see Appendix A for a detailed derivation)
\begin{align}\label{c_2_7}
W(n)=&P[W(n-1)+\eta g(e(n))(d(n)- \nonumber\\
  &W^{T}(n-1)X(n))X(n)]+Q
\end{align}
where $g(e(n))$ is a nonlinear function of $e(n)$, given by
\begin{align}\label{c_2_8}
g(e(n))=\exp\left(-\frac{e^{2}(n)}{2\sigma^{2}}\right)
\end{align}
The above algorithm is referred to as the CMCC algorithm, whose pseudocodes are presented in Table \ref{cmcc algorithm}.

\subsection{Computational Complexity}
The computational complexity of the proposed CMCC algorithm and other constrained adaptive algorithms-CLMS, CAP and CRLS, in terms of the total number of required additions and multiplications at each iteration, are shown in Table \ref{computation complexity}, where $\Gamma_{g}$ is a constant associated with the complexity of the nonlinear function $g(e(n))$. Obviously, the computational complexity of these algorithms are $O(M^{2})$. When $\Gamma_{g}$ is small, it can be seen that the proposed algorithm has lower computational cost than CRLS due to calculating the covariance matrix $\textbf{R}$ per iteration for CRLS, also has lower computational cost than CAP (especially when the sliding window length $L$ is large). Generally speaking, the computational complexity of CMCC is almost the same as that of the CLMS.
\begin{table}[!t]
\renewcommand{\arraystretch}{1.3}
\caption{CMCC Algorithm}
\label{cmcc algorithm}
\centering
\begin{tabular}{l}
    \hline \\
    Parameters: \;\;\;\;$\eta$, $\sigma$, $C$ and $f$ \\ \\
    Initialization :\;$P=\textbf{I}_{M}-C(C^{T}C)^{-1}C^{T}$ \\
    \;\;\;\;\;\;\;\;\;\;\;\;\;\;\;\;\;\;\;\;\;$Q=C(C^{T}C)^{-1}f$ \\
    \;\;\;\;\;\;\;\;\;\;\;\;\;\;\;\;\;\;\;\;\;$W(0)=Q$ \\ \\
    \hline
    \\
    Update: \;\;\;\;\;\;\; $y(n)=W^{T}(n-1)X(n)$\\
    \;\;\;\;\;\;\;\;\;\;\;\;\;\;\;\;\;\;\;\;\;$e(n)=d(n)-y(n)$ \\
    \;\;\;\;\;\;\;\;\;\;\;\;\;\;\;\;\;\;\;\;\;$W(n)=P\left[W(n-1)+ \eta g(e(n))e(n)X(n)\right]+Q$  \\ \\
    \hline
\end{tabular}
\end{table}
\begin{table}[!t]
\centering
\caption{Computational Complexity of CMCC, CLMS, CAP and CRLS}\label{computational complexity}
\label{computation complexity}
\begin{tabular}{c|c}
  \hline\hline
   Algorithm & Computational Complexity  \\
  \hline
   CMCC & $2M^{2}+5M+\Gamma_{g}$  \\
   CLMS & $2M^{2}+5M+1$  \\
   CAP & $2M^{2}+(2L+3)M+1$  \\
   CRLS & $7M^{2}+(6K^{2}+9K+5)M+3K$  \\
  \hline
\end{tabular}
\end{table}
%
%
\section{Convergence Analysis}
In this section, we analyze the mean square convergence behaviors of the proposed CMCC algorithm. First, we give the following assumptions:
\begin{enumerate}
  \item The input sequence $\left\{X(n)\right\}$ is independent multivariate Gaussian, with zero-mean and the positive-definite covariance matrix of the input sequence $\textbf{R}=E[X(n)X^{T}(n)]$.
  \item The noise $\left\{\upsilon(n)\right\}$ is zero-mean, independent, identically distributed, and independent of any other signals in the system.
  \item The error nonlinearity $g(e(n))$ is asymptotically uncorrelated with $\left\{X(n)X^{T}(n)\right\}$ at steady-state.
  \item The filter is long enough such that the a priori error $e_{a}(n)=\left(W^{\ast}-W(n)\right)^{T}X(n)$ is zero-mean Gaussian.
\end{enumerate} \par
 The independence assumptions 1) and 2) are common in the literature of adaptive filtering \cite{36,37,38,39}. When the filter is long enough, assumption 3) will become realistic and valid. Assumption 4) is reasonable by the central limit theorem, and also remains valid in the whole stage of adaptation (see \cite{18,38} for more detailed explanation about assumptions 3) and 4)).
\subsection{Mean Square Stability}
Let us define the weight error vector:
\begin{align}\label{t_1_1}
\tilde{W}(n)=W(n)-W_{opt}
\end{align}
where $W_{opt}$ stands for the optimal solution of the CMCC optimization problem under the above assumptions, given by (see Appendix B for the detailed derivation)
\begin{align}\label{t_1_2}
{W}_{opt}=W^{*}+\textbf{R}_{g}^{-1}C\left(C^{T}\textbf{R}_{g}^{-1}C\right)^{-1} \left(f-C^{T}W^{*}\right)
\end{align}
where $\textbf{R}_{g}=E\left[g(e(n))X(n)X^{T}(n)\right]$ denotes a weighted autocorrelation matrix of the input vector. We also define
\begin{align}\label{t_1_3}
\varepsilon_{w}=W^{*}-{W}_{opt}
\end{align}
Substituting (\ref{c_2_1}), (\ref{t_1_1}) and (\ref{t_1_3}) into (\ref{c_2_7}) yields
\begin{align}\label{t_1_4}
\tilde{W}(n)=&P\left[W(n-1)+\eta g(e(n))(d(n)-\right. \nonumber \\
             &\left. W^{T}(n-1)X(n))X(n)\right]+Q-W_{opt}  \nonumber \\
           =&P\left[\textbf{I}_{M}-\eta g(e(n))X(n)X^{T}(n)\right]\tilde{W}(n-1)+ \nonumber \\
            &\eta g(e(n))\upsilon (n)PX(n)+ \eta g(e(n))PX(n)\times  \nonumber \\
            & X^{T}(n)\varepsilon_{w}+P{W}_{opt}-{W}_{opt}+Q
\end{align}
Due to $P{W}_{opt}-{W}_{opt}+Q=\textbf{0}_{M\times1}$ (Here $\textbf{0}_{M\times1}$ denotes the $M\times1$ zero vector), we can rewrite (\ref{t_1_4}) as
\begin{align}\label{t_1_5}
\tilde{W}(n)=&P\left[\textbf{I}_{M}-\eta g(e(n))X(n)X^{T}(n)\right]\times \nonumber \\
           & \tilde{W}(n-1)+\eta g(e(n))\upsilon (n)PX(n)+ \nonumber \\
           &\eta g(e(n))PX(n)X^{T}(n)\varepsilon_{w}
\end{align}
Note that matrix $P$ is idempotent, namely $P=P^{2}$ and $P=P^{T}$. Multiplying both sides of (\ref{t_1_5}) by $P$ and after some straightforward matrix manipulations, we can obtain
\begin{align}\label{t_1_6}
P\tilde{W}(n)=\tilde{W}(n)
\end{align}
Combining (\ref{t_1_5}) and (\ref{t_1_6}), we have
\begin{align}\label{t_1_7}
\tilde{W}(n)=&P\tilde{W}(n-1)-\eta g(e(n))PX(n)X^{T}(n)\tilde{W}(n-1)+\nonumber \\
            &\eta g(e(n))\upsilon (n)PX(n)+\eta g(e(n))PX(n)X^{T}(n)\varepsilon_{w} \nonumber \\
            =&\left(\textbf{I}_{M}-\eta g(e(n))PX(n)X^{T}(n)P\right)\tilde{W}(n-1)+ \nonumber \\
            &\eta g(e(n))\upsilon (n)PX(n)+\eta g(e(n))PX(n)X^{T}(n)\varepsilon_{w}
\end{align}
Under assumptions 1), 2) and 3), taking the expectations of the squared Euclidean norms of both sides of (\ref{t_1_7}) leads to the following energy conservation relation:
\begin{align}\label{t_1_8}
&E\left[\|\tilde{W}(n)\|^{2}\right]=E\left[\|\tilde{W}(n-1)\|^{2}_{\textbf{H}}\right]+\eta^{2}E\left[g^{2}(e(n))\right]\times \nonumber \\
&\qquad\qquad E\left[\upsilon^{2}(n)\right]E\left[X^{T}(n)PX(n)\right]+\eta^{2}E\left[g^{2}(e(n))\right]   \nonumber \\
&\qquad\qquad\times\varepsilon^{T}_{w}E\left[X(n)X^{T}(n)PX(n)X^{T}(n)\right]\varepsilon_{w}
\end{align}
where $E\left[\|\tilde{W}(n)\|^{2}\right]$ is called the \emph{weight error power} (WEP) at iteration $n$ , $\|\tilde{W}(n-1)\|^{2}_{\textbf{H}}=\tilde{W}^{T}(n-1)\textbf{H}\tilde{W}(n-1)$ , and
\begin{align*}
\textbf{H}=&\textbf{I}_{M}-2\eta E\left[g(e(n))\right]P\textbf{R}P+\eta^{2}E\left[g^{2}(e(n))\right]\times \nonumber \\ &PE[X(n)X^{T}(n)PX(n)X^{T}(n)]P
\end{align*}
Since $P=P^{2}$, we derive
\begin{align}\label{t_1_9}
E\left[X^{T}(n)PX(n)\right]=&E\left[X^{T}(n)PPX(n)\right] \nonumber \\
=&tr\left\{P\textbf{R}P\right\} \nonumber \\
=&tr\left\{\Upsilon\right\}
\end{align}
where $tr\left\{\cdot\right\}$ stands for the \emph{trace operator}, and $\Upsilon=P\textbf{R}P$.
According to the Isserlis' theorem \cite{40} for Gaussian vectors $\hbar_{1}$, $\hbar_{2}$, $\hbar_{3}$ and $\hbar_{4}$, we have
\begin{align}\label{t_1_10}
E\left[\hbar_{1}\hbar_{2}^{T}\hbar_{3}\hbar_{4}^{T}\right]=&E\left[\hbar_{1}\hbar_{2}^{T}\right]E\left[\hbar_{3}\hbar_{4}^{T}\right]+
E\left[\hbar_{1}\hbar_{3}^{T}\right]E\left[\hbar_{2}\hbar_{4}^{T}\right] \nonumber \\
+&E\left[\hbar_{1}\hbar_{4}^{T}\right]E\left[\hbar_{2}^{T}\hbar_{3}\right]
\end{align}
With $\hbar_{1}=X(n)$, $\hbar_{2}=X(n)$, $\hbar_{3}=PX(n)$ and $\hbar_{4}=X(n)$, we obtain
\begin{align}\label{t_1_11}
&E\left[X(n)X^{T}(n)PX(n)X^{T}(n)\right]  \nonumber \\
&\qquad\qquad =\textbf{R}P\textbf{R}+\textbf{R}P\textbf{R}+E\left[X^{T}(n)PX(n)\right]\textbf{R} \nonumber \\
&\qquad\qquad =tr\left\{\Upsilon\right\}\textbf{R}+2\textbf{R}P\textbf{R}
\end{align}
Since $P\textbf{R}\varepsilon_{w}=\textbf{0}_{M\times1}$, Substituting (\ref{t_1_9}) and (\ref{t_1_11}) into (\ref{t_1_8}), we get
\begin{align}\label{t_1_12}
E\left[\|\tilde{W}(n)\|^{2}\right]=&E\left[\|\tilde{W}(n-1)\|^{2}_{\textbf{H}}\right]+\eta^{2}E\left[g^{2}(e(n))\right]\times \nonumber\\
&tr\left\{\Upsilon\right\}\left(\varepsilon^{T}_{w}\textbf{R}\varepsilon_{w}+E\left[\upsilon^{2}(n)\right]\right)
\end{align}
and
\begin{align*}
\textbf{H}=&\textbf{I}_{M}-2\eta E\left[g(e(n))\right]P\textbf{R}P+\eta^{2}E\left[g^{2}(e(n))\right]\times \nonumber \\ &(tr\left\{\Upsilon\right\}P\textbf{R}P+2P\textbf{R}P\textbf{R}P)
\end{align*}\par
Let $\lambda_{i}\left(i=1,\ldots,M-K\right)$ be the eigenvalues of the matrix $\Upsilon$. A sufficient condition for the mean square stability can be obtained as \cite{5,24,37}
\begin{align}\label{t_1_13}
&\left|1-2\eta E\left[g(e(n))\right]\lambda_{i}+\eta^{2}E\left[g^{2}(e(n))\right]tr\left\{\Upsilon\right\}\lambda_{i}+\right. \nonumber\\
&\qquad\qquad\qquad\qquad\qquad\qquad \left. 2\eta^{2}E\left[g^{2}(e(n))\right]\lambda_{i}^{2}\right|<1  \nonumber\\
&\qquad\qquad\qquad\qquad i=1,\ldots,M-K
\end{align}
After some simple manipulations, we have
\begin{align}\label{t_1_14}
0<\eta<\frac{2E\left[g(e(n))\right]}{\left[2\lambda_{max}+tr\left\{\Upsilon\right\}\right]E\left[g^{2}(e(n))\right]}
\end{align}
where $\lambda_{max}$ denotes the largest eigenvalue of the matrix $\Upsilon$.
Due to $E\left[g(e(n))\right]\geq E\left[g^{2}(e(n))\right]>0$, one can obtain a stronger condition to guarantee the mean square stability:
\begin{align}\label{t_1_15}
0<\eta\leq\frac{2}{2\lambda_{max}+tr\left\{\Upsilon\right\}}
\end{align}
\textbf{Remark:} Since we only derive (\ref{t_1_14}) and (\ref{t_1_15}) under the steady-state assumption, we cannot solve the problem of how to select the best step-size for a specific application. However, the condition provides a possible range for choosing a step-size for CMCC algorithm.
\subsection{Steady-state mean square deviation (MSD)}
 Assume that $\textbf{T}$ is an arbitrary symmetric nonnegative definite matrix. Under assumptions 1), 2) and 3) , one can derive the following relation by taking the expectations of the squared -weighted Euclidean norms of both sides of (\ref{t_1_7}):
\begin{align}\label{t_2_1}
E\left[\|\tilde{W}(n)\|_{\textbf{T}}^{2}\right]=&E\left[\|\tilde{W}(n-1)\|_{\textbf{U}}^{2}\right]+\eta^{2}E\left[g^{2}(e(n))\right]\times \nonumber \\
&E\left[\upsilon^{2}(n)\right]E\left[X^{T}(n)P\textbf{T}PX(n)\right]+\eta^{2}\times   \nonumber \\
&E\left[g^{2}(e(n))\right]\varepsilon^{T}_{w}E\left[X(n)X^{T}(n)\times\right. \nonumber \\
&\left.P\textbf{T}PX(n)X^{T}(n)\right]\varepsilon_{w}
\end{align}
in which
\begin{align}\label{t_2_2}
\textbf{U}=&E\left[(\textbf{I}_{M}-\eta g(e(n))X(n)X^{T}(n))P\textbf{T}P\times\right.  \nonumber \\
           &\left.\left(\textbf{I}_{M}-\eta g(e(n))X(n)X^{T}(n)\right)\right]          \nonumber \\
          =&P\textbf{T}P-\eta E\left[g(e(n))\right]\textbf{R}P\textbf{T}P-\eta\times \nonumber\\
           &E\left[g(e(n))\right]P\textbf{T}P\textbf{R}+\eta^{2}E\left[g^{2}(e(n))\right] \times \nonumber\\
           & E\left[X(n)X^{T}(n)P\textbf{T}PX(n)X^{T}(n)\right]
\end{align}
In the same way as for (\ref{t_1_9}) and (\ref{t_1_11}), we derive
\begin{align}\label{t_2_3}
E\left[X^{T}(n)P\textbf{T}PX(n)\right]=tr\left\{\textbf{T}\Upsilon\right\}
\end{align}
%
\begin{align}\label{t_2_4}
&E\left[X(n)X^{T}(n)P\textbf{T}PX(n)X^{T}(n)\right] \nonumber \\
&\qquad\qquad\qquad =tr\left\{\textbf{T}\Upsilon\right\}\textbf{R}+2\textbf{R}P\textbf{T}P\textbf{R}
\end{align}
Thus we can rewrite (\ref{t_2_2}) as
\begin{align}\label{t_2_5}
\textbf{U}=&\left(\textbf{I}_{M}-\eta E\left[g(e(n))\right]\textbf{R}\right)P\textbf{T}P \left(\textbf{I}_{M}-\eta E\left[g(e(n))\right]\textbf{R}\right) + \nonumber\\
                &\eta^{2}E\left[g^{2}(e(n))\right]tr\left\{\textbf{T}\Upsilon\right\}\textbf{R}+ 2\eta^{2}E\left[g^{2}(e(n))\right]\times \nonumber\\
                &\textbf{R}P\textbf{T}P\textbf{R}-\eta^{2}E^{2}\left[g(e(n))\right]\textbf{R}P\textbf{T}P\textbf{R}
\end{align}
From \cite{41}, some useful properties can be obtained, that is,
\begin{align*}
  \textbf{vec\{BCD\}}=\left(\textbf{D}^{T}\otimes \textbf{B}\right)\textbf{vec\{C\}}
\end{align*}
and
\begin{align*}
  tr\left\{\textbf{B}^{T}\textbf{C}\right\}=\textbf{vec}^{T}\left\{\textbf{C}\right\}\textbf{vec}\left\{\textbf{B}\right\}
\end{align*}
where $\textbf{vec}\left\{\cdot\right\}$ denotes the vectorization operator, $\otimes$ stands for the Kronecker product. With the vectorization and the above properties, we have
\begin{align}\label{t_2_6}
\textbf{vec}\left\{\textbf{U}\right\}=\textbf{F}\textbf{t}
\end{align}
where
\begin{align*}
  \textbf{F}=&\left(\textbf{I}_{M}-\eta E\left[g(e(n))\right]\textbf{R}\right)P\otimes\left(\textbf{I}_{M}-\eta E\left[g(e(n))\right]\textbf{R}\right)P \nonumber\\
                &+2\eta^{2}E\left[g^{2}(e(n))\right]\left(\textbf{R}P\otimes\textbf{R}P\right)+ \eta^{2}E\left[g^{2}(e(n))\right]\times  \nonumber\\
                &\textbf{vec}\left\{\textbf{R}\right\}\textbf{vec}\left\{\Upsilon\right\}-\eta^{2}E^{2}\left[g(e(n))\right]\left(\textbf{R}P\otimes\textbf{R}P\right)
\end{align*}
and $\textbf{t}=\textbf{vec}\left\{{\textbf{T}}\right\}$ . Combining (\ref{t_2_3}), (\ref{t_2_4}) and (\ref{t_2_6}), we can rewrite (\ref{t_2_1}) as
\begin{align}\label{t_2_7}
E\left[\|\tilde{W}(n)\|_{\textbf{t}}^{2}\right]=&E\left[\|\tilde{W}(n-1)\|_{\textbf{Ft}}^{2}\right]+\eta^{2}E\left[g^{2}(e(n))\right]\times \nonumber \\
&\left(\varepsilon^{T}_{w}\textbf{R}\varepsilon_{w}+E\left[\upsilon^{2}(n)\right]\right)\textbf{vec}^{T}\left\{\Upsilon\right\}\textbf{t}
\end{align}
We define the steady-state MSD as follows:
\begin{align}\label{t_2_8}
S=\lim\limits_{n\rightarrow\infty}E\left[\|\tilde{W}(n)\|^{2}\right]
\end{align}
Assume that the filter is stable and achieves the steady-state, i.e. $\lim\limits_{n\rightarrow\infty}E\left[\|\tilde{W}(n)\|^{2}\right]=\lim\limits_{n\rightarrow\infty}E\left[\|\tilde{W}(n-1)\|^{2}\right]$. By (\ref{t_2_7}), we have
\begin{align}\label{t_2_9}
\lim\limits_{n\rightarrow\infty}E\left[\|\tilde{W}(n)\|_{\left(\textbf{I}_{M^{2}}-\textbf{F}\right)\textbf{t}}^{2}\right]=&
  \lim\limits_{n\rightarrow\infty}
  \eta^{2}E\left[g^{2}(e(n))\right]\times \nonumber \\
  &\left(\varepsilon^{T}_{w}\textbf{R}\varepsilon_{w}+E\left[\upsilon^{2}(n)\right]\right)\times\nonumber \\
  &\textbf{vec}^{T}\left\{\Upsilon\right\}\textbf{t}
\end{align}
Therefore, by selecting an appropriate $\textbf{t}=\left(\textbf{I}_{M^{2}}-\textbf{F}\right)^{-1}\textbf{vec}\{\textbf{I}_{M}\}$, we can obtain
\begin{align}\label{t_2_10}
S=&\eta^{2}\left(\varepsilon^{T}_{w}\textbf{R}\varepsilon_{w}+E\left[\upsilon^{2}(n)\right]\right)\textbf{vec}^{T}\left\{\Upsilon\right\}\times \nonumber \\
&\lim\limits_{n\rightarrow\infty}\left(\textbf{I}_{M^{2}}-\textbf{F}\right)^{-1}\textbf{vec}\{\textbf{I}_{M}\}E\left[g^{2}(e(n))\right]
\end{align}
Based on assumption 3), we can rewrite (\ref{t_1_2}) as following:
\begin{align}\label{t_2_11}
{W}_{opt}=&W^{*}+\textbf{R}^{-1}C\left(C^{T}\textbf{R}^{-1}C\right)^{-1}\left(f-C^{T}W^{*}\right)
\end{align}
and accordingly
\begin{align}\label{t_2_12}
\varepsilon_{w}=\textbf{R}^{-1}C\left(C^{T}\textbf{R}^{-1}C\right)^{-1}\left(C^{T}W^{*}-f\right)
\end{align}
In order to obtain the theoretical value of the steady-state MSD, we also need to evaluate the values of $\lim\limits_{n\rightarrow\infty}E\left[g(e(n))\right]$ and $\lim\limits_{n\rightarrow\infty}E\left[g^{2}(e(n))\right]$. We consider two cases below:
\begin{enumerate}
  \item If $\upsilon(n)$ is zero-mean Gaussian distributed with variance $\sigma_{\upsilon}^{2}$, then
    \begin{align}\label{t_2_13}
        \lim\limits_{n\rightarrow\infty}E\left[g(e(n))\right]\approx\frac{\sigma}{\sqrt{\sigma^{2}+\varepsilon^{T}_{w}\textbf{R}\varepsilon_{w}+\sigma_{\upsilon}^{2}}}
    \end{align}
    %
    \begin{align}\label{t_2_14}
        \lim\limits_{n\rightarrow\infty}E\left[g^{2}(e(n))\right]\approx\frac{\sigma}{\sqrt{\sigma^{2}+2\varepsilon^{T}_{w}\textbf{R}\varepsilon_{w}+2\sigma_{\upsilon}^{2}}}
    \end{align}
    Thus
    \begin{align}\label{t_2_15}
        S\approx &\eta^{2}\left(\varepsilon^{T}_{w}\textbf{R}\varepsilon_{w}+\sigma_{\upsilon}^{2}\right) \textbf{vec}^{T}\left\{\Upsilon\right\}\left(\textbf{I}_{M^{2}}-\textbf{F}\right)^{-1}\times \nonumber \\
        &\quad\textbf{vec}\{\textbf{I}_{M}\}\frac{\sigma}{\sqrt{\sigma^{2}+2\varepsilon^{T}_{w}\textbf{R}\varepsilon_{w}+2\sigma_{\upsilon}^{2}}}
    \end{align}
    \item If $\upsilon(n)$ is non-Gaussian, then by Taylor expansion we have
    \begin{align}\label{t_2_16}
        \lim\limits_{n\rightarrow\infty}E\left[g(e(n))\right]\approx E\left[\exp\left(-\frac{\upsilon^{2}(n)}{2\sigma^{2}}\right)\right]+\frac{1}{2}\varepsilon^{T}_{w}\textbf{R}\varepsilon_{w}\times \nonumber \\ E\left[\left(\frac{\upsilon^{2}(n)}{\sigma^{4}}-\frac{1}{\sigma^{2}}\right)\exp\left(-\frac{\upsilon^{2}(n)}{2\sigma^{2}}\right)\right]
    \end{align}
    %
    \begin{align}\label{t_2_17}
        \lim\limits_{n\rightarrow\infty}E\left[g^{2}(e(n))\right]\approx E\left[\exp\left(-\frac{\upsilon^{2}(n)}{\sigma^{2}}\right)\right]+\varepsilon^{T}_{w}\textbf{R}\varepsilon_{w}\times \nonumber \\ E\left[\left(\frac{2\upsilon^{2}(n)}{\sigma^{4}}-\frac{1}{\sigma^{2}}\right)\exp\left(-\frac{\upsilon^{2}(n)}{\sigma^{2}}\right)\right]
    \end{align}
    It follows that
  \begin{align}\label{t_2_18}
        S\approx& \eta^{2}\left(\varepsilon^{T}_{w}\textbf{R}\varepsilon_{w}+E\left[\upsilon^{2}(n)\right]\right)\textbf{vec}^{T}\left\{\Upsilon\right\} \left(\textbf{I}_{M^{2}}-\textbf{F}\right)^{-1}\times \nonumber\\
        &\textbf{vec}\{\textbf{I}_{M}\}\left(E\left[\exp\left(-\frac{\upsilon^{2}(n)}{\sigma^{2}}\right)\right]+\right.
        \varepsilon^{T}_{w}\textbf{R}\varepsilon_{w}\times    \nonumber \\
        &\left.E\left[\left(\frac{2\upsilon^{2}(n)}{\sigma^{4}}-\frac{1}{\sigma^{2}}\right)\exp\left(-\frac{\upsilon^{2}(n)}{\sigma^{2}}\right)\right]\right)
    \end{align}
\end{enumerate}
\textbf{Remark:} It is worth noting that (\ref{t_2_15}) and (\ref{t_2_18}) have been derived by using the approximation $W(n)\approx W_{opt}$ at the steady state. In addition, the theoretical value for non-Gaussian noise case has been derived by taking the Taylor expansion of $g(e(n))$ around $\upsilon(n)$ and omitting the higher-order terms. If the noise power is very large, the approximation is not accurate and hence, the derived values at steady state may deviate seriously from the actual results. The detailed derivations for (\ref{t_2_13}) to (\ref{t_2_18}) can be found in Appendix C.

%
%
\section{Simulation Results}
In this section, we present simulation results to confirm the theoretical conclusions drawn in the previous section, and illustrate the superior performance of the proposed CMCC algorithm compared with the traditional CLMS algorithm \cite{5}, CAP algorithm \cite{11} and CRLS algorithm \cite{12}. The selection of kernel bandwidth is also discussed in the end.
\subsection{Non-Gaussian Noise Models}
Generally speaking, the non-Gaussian noise distributions can be divided into two categories: light-tailed (e.g. binary, uniform, etc.) and heavy-tailed (e.g. Laplace, Cauchy, mixed Gaussian, alpha-stable, etc.) distributions \cite{31, 38, 39, 42, 43}. In the following experiments, five common non-Gaussian noise models including binary noise, Laplace noise, Cauchy noise, Mixed Gaussian noise, and alpha-stable noise, are selected for performance evaluation. Descriptions of these non-Gaussian noises are as following:
\begin{enumerate}
  \item Binary noise model: Standard binary noise takes the values of either $\upsilon=1$ or $\upsilon=-1$, with probability mass function $Pr\{\upsilon=1\}=Pr\{\upsilon=-1\}=0.5$.
    %
  \item Laplace noise model: The Laplace noise is distributed with probability density function (PDF):
    \begin{align}\label{Laplace noise}
        p(\upsilon)=\frac{1}{2}\exp^{-|\upsilon|}
    \end{align}
    \item Cauchy noise model: The PDF of the Cauchy noise is
    \begin{align}\label{Cauchy noise}
        p(\upsilon)=\frac{1}{\pi(1+\upsilon^{2})}
    \end{align}
    \item Mixed Gaussian noise model: The mixed Gaussian noise model is given by:
    \begin{align}\label{mixed Gaussian noise}
    (1-\theta)\mathcal{N}\left(\lambda_{1}, \upsilon^{2}_{1}\right)+\theta \mathcal{N}\left(\lambda_{2}, \upsilon^{2}_{2}\right)
    \end{align}
    where $\mathcal{N}\left(\lambda_{i}, \upsilon^{2}_{i}\right)(i=1,2)$ denote the Gaussian distributions with mean values $\lambda_{i}$ and variances $\upsilon^{2}_{i}$, and $\theta$ is the mixture coefficient. Usually one can set $\theta$ to a small value and $\upsilon^{2}_{2}\gg\upsilon^{2}_{1}$  to represent the impulsive noises (or large outliers). Therefore, we define the mixed Gaussian noise parameter vector as  $V_{mix}=\left(\lambda_{1},\lambda_{2},\upsilon^{2}_{1},\upsilon^{2}_{2},\theta\right)$.
    \item Alpha-stable noise model: The characteristic function of the alpha-stable noise is defined as:
    \begin{align}\label{alpha-stable noise}
    \psi(t)=\exp\{j\delta t-\gamma|t|^\alpha[1 + j\beta sgn(t)S(t,\alpha )]\}
    \end{align}
    in which
    \begin{align}
    S(t,\alpha )=\left\{
        \begin{array}{cc}
          \tan(\frac{\alpha\pi}{2})  & \emph{if}\;\;\; \alpha\neq1\\
          \frac{2}{\pi}\log|t|       & \emph{if}\;\;\; \alpha=1
        \end{array}\right.
    \end{align}
    From (\ref{alpha-stable noise}), one can observe that a stable distribution is completely determined by four parameters: 1) the characteristic factor $\alpha$; 2) the symmetry parameter $\beta$; 3) the dispersion parameter $\gamma$; 4) the location parameter $\delta$. So we define the alpha-stable noise parameter vector as $V_{alpha}=(\alpha,\beta,\gamma,\delta)$.
\end{enumerate}

It is worth mentioning that, in the case of $\alpha=2$, the alpha-stable distribution coincides with the Gaussian distribution, while $\alpha=1, \delta=0$ is the same as the Cauchy  distribution.
\subsection{Validation of Steady-state MSD}
\begin{figure}[!t]
\centering
\subfloat{\includegraphics[width=2.5in]{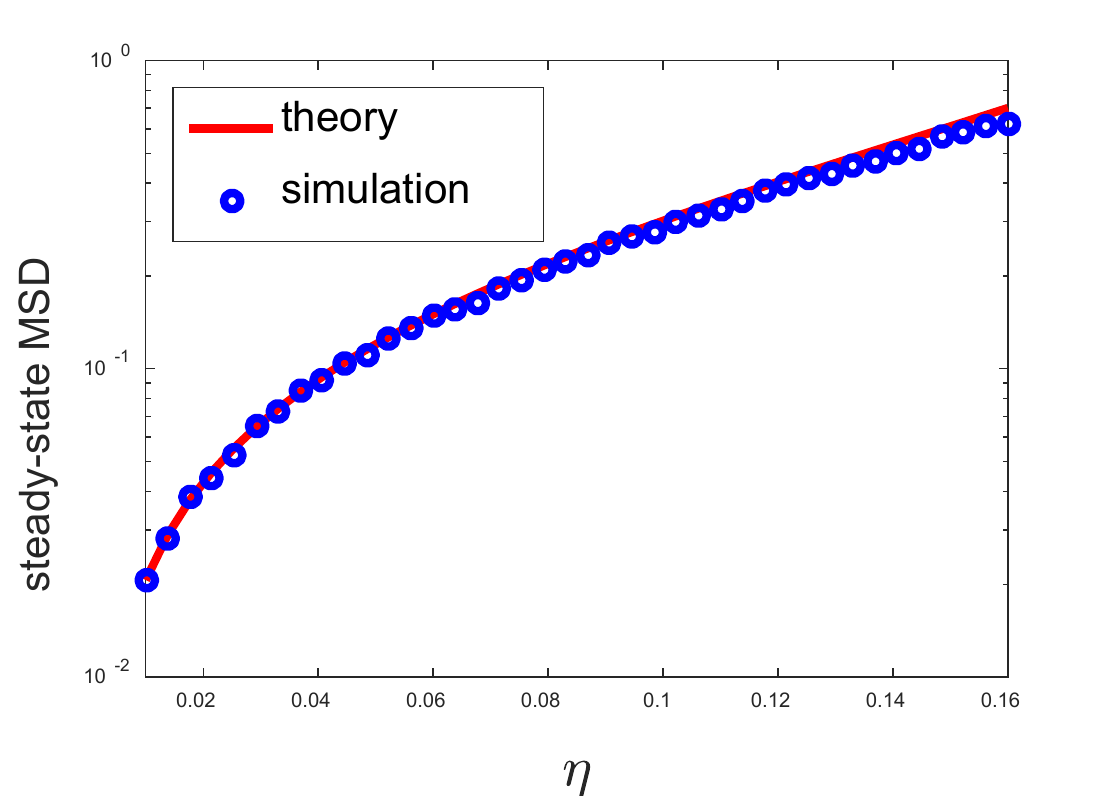}
\label{pt-gaussian-step}}
\hfil
\\ (a)\\
\subfloat{\includegraphics[width=2.5in]{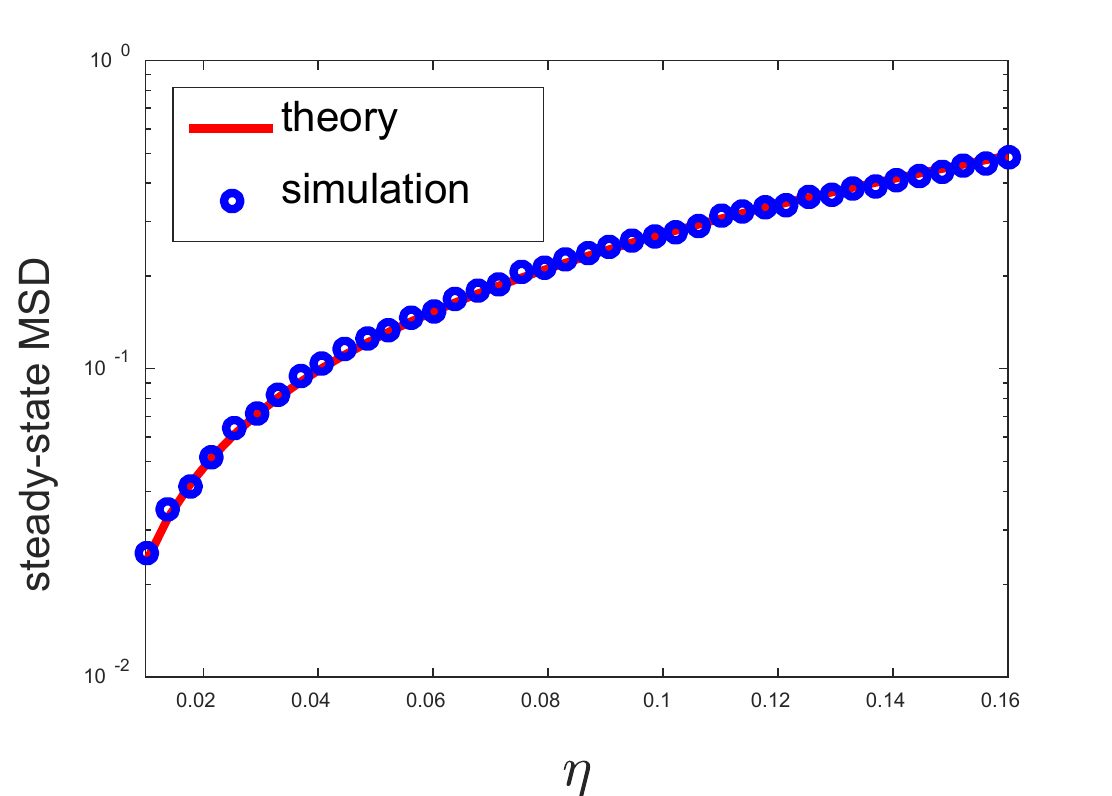}
\label{pt-binary-step}}
\hfil
\\ (b)\\
\subfloat{\includegraphics[width=2.5in]{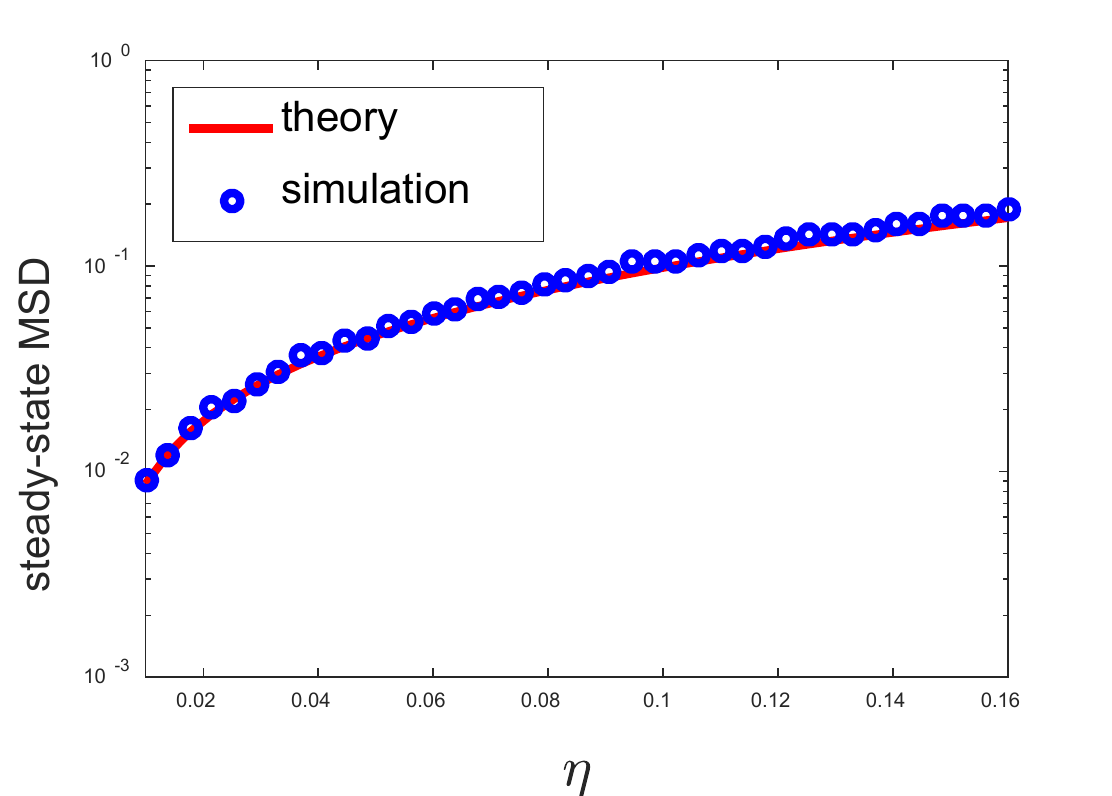}
\label{pt-laplace-step}}
\\ (c)\\
\caption{Theoretical and simulated steady-state MSDs with different step-sizes $\eta$: (a) Gaussian noise ($\sigma=8.0$, $\sigma^{2}_{\upsilon}=0.81$); (b) Binary noise ($\sigma=2.0$, $\sigma^{2}_{\upsilon}=1.0$); (c) Laplace noise ($\sigma=1.0$, $\sigma^{2}_{\upsilon}=1.0$).}
\label{pt-step}
\end{figure}

\begin{figure}[!t]
\centering
\subfloat{\includegraphics[width=2.5in]{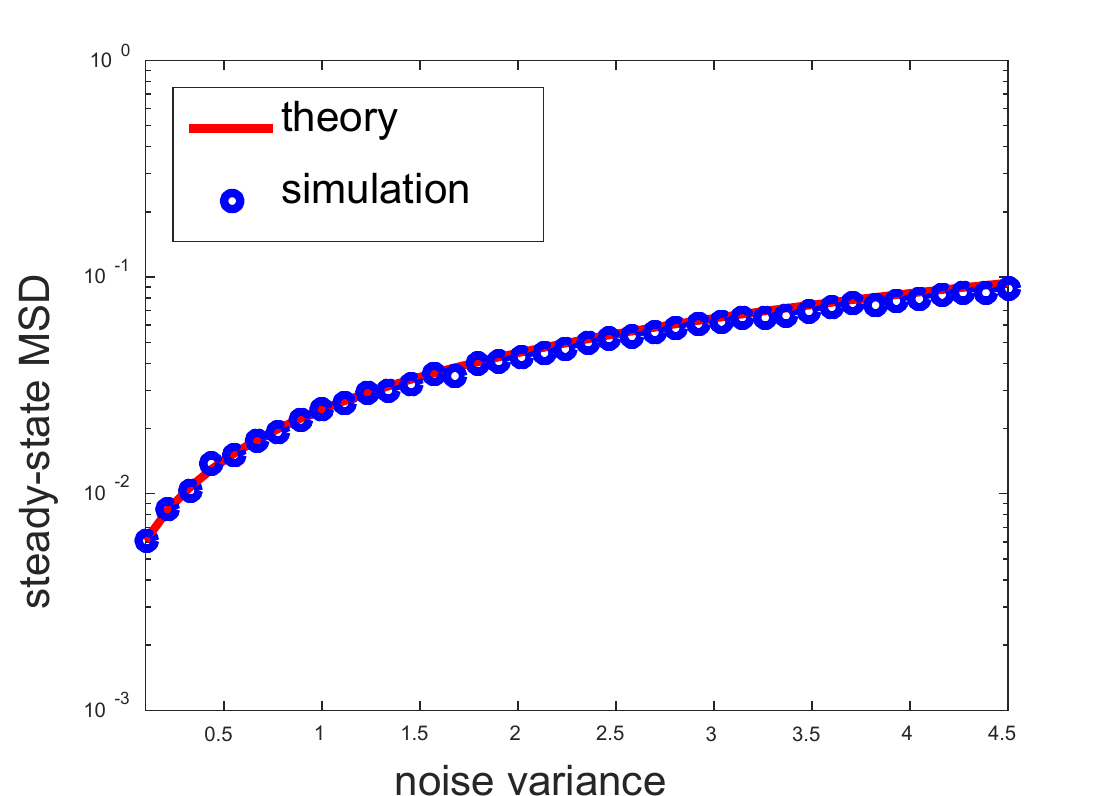}
\label{pt-gaussian-variance}}
\hfil
\\ (a)\\
\subfloat{\includegraphics[width=2.5in]{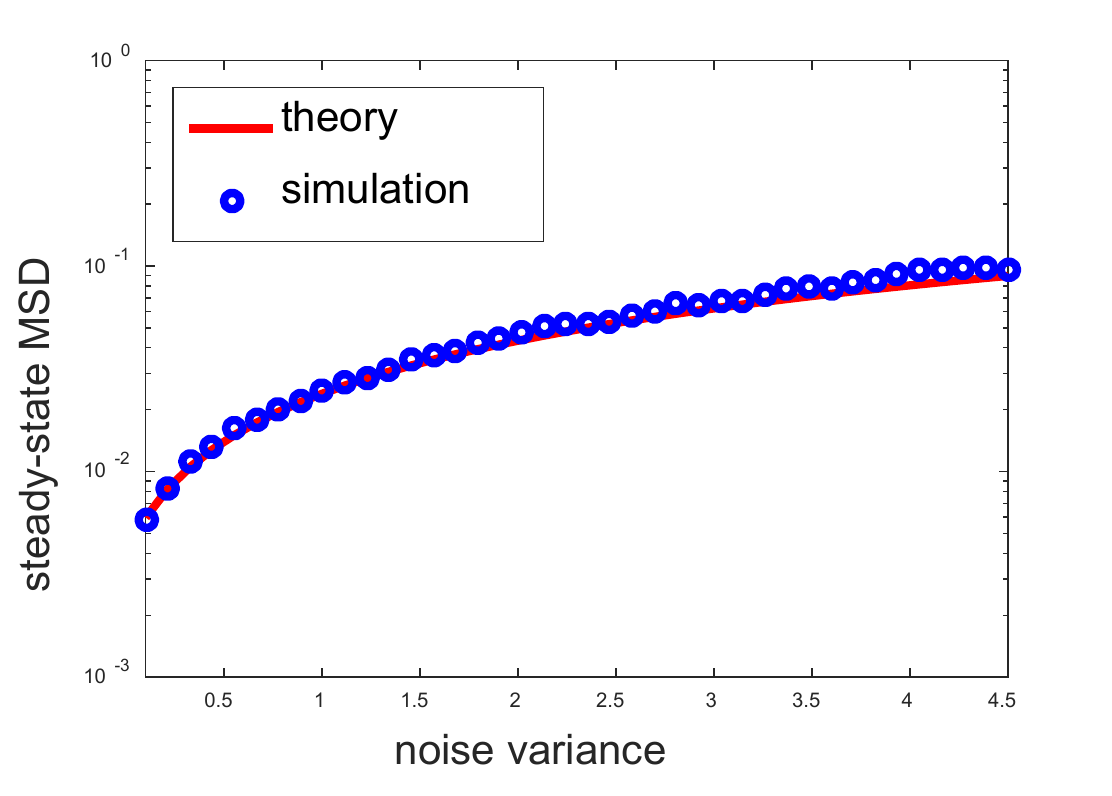}
\label{pt-binary-variance}}
\hfil
\\ (b)\\
\subfloat{\includegraphics[width=2.5in]{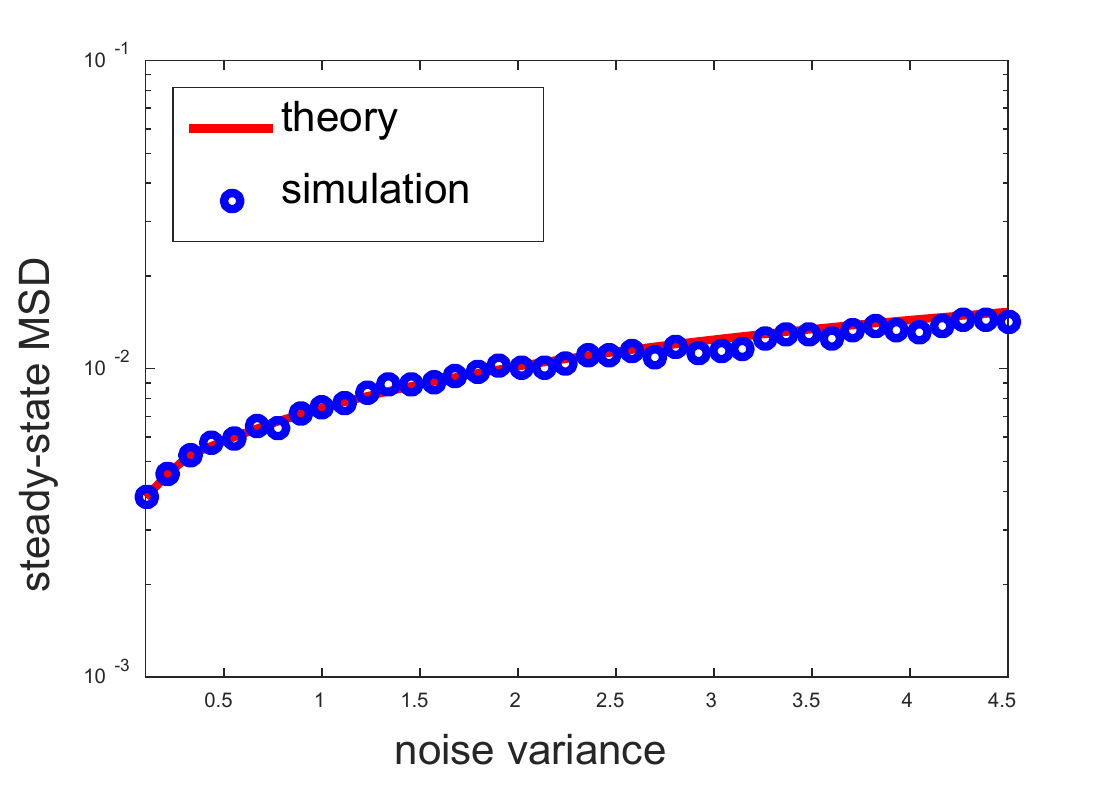}
\label{pt-laplace-variance}}
\\ (c)\\
\caption{Theoretical and simulated steady-state MSDs with different noise variance $\sigma^{2}_{\upsilon}$: (a) Gaussian noise ($\eta=0.01$, $\sigma=8.0$); (b) Binary noise ($\eta=0.01$, $\sigma=6.0$); (c) Laplace noise ($\eta=0.01$, $\sigma=0.8$).}
\label{pt-variance}
\end{figure}

In this experiment, we show the values of the theoretical and simulated steady-state MSDs of the CMCC in a linear channel with weight vector $(M=7)$
\begin{align}
W^{*}=&[0.332, -0.040, -0.094, 0.717, \nonumber \\
&\qquad\qquad\qquad\qquad -0.652, -0.072, 0.580]^{T}
\end{align}
Assume that $K=3$, $C$ is full-rank, and the input covariance matrix $\textbf{R}$ is positive-definite with $tr\left\{\textbf{R}\right\}=M$ \cite{13}. The input vectors are zero-mean multrivate Gaussian, and the disturbance noises considered include Gaussian noise, binary noise (light-tailed disturbance) and Laplace noise (heavy-tailed disturbance). Fig. \ref{pt-step} shows the theoretical and simulated steady-state MSDs with different step-sizes, and Fig. \ref{pt-variance} presents the theoretical and simulated steady-state MSDs with different noise variances. If not mentioned otherwise, simulation results are averaged over 500 independent Monte Carlo runs, and in each simulation, 5000 iterations are run to ensure the algorithms to reach the steady state, and the steady-state MSDs are obtained as averages over the last 200 iterations. Evidently,  the steady-state MSDs are increasing with the step-size and noise variances increasing. In addition, the steady-state MSDs obtained from simulations match well with those theoretical results (computed by (\ref{t_2_13}) for Gaussian noise and (\ref{t_2_18}) for Non-Gaussian noise).
\begin{figure*}[!t]
\centering
\subfloat{\includegraphics[width=2.5in]{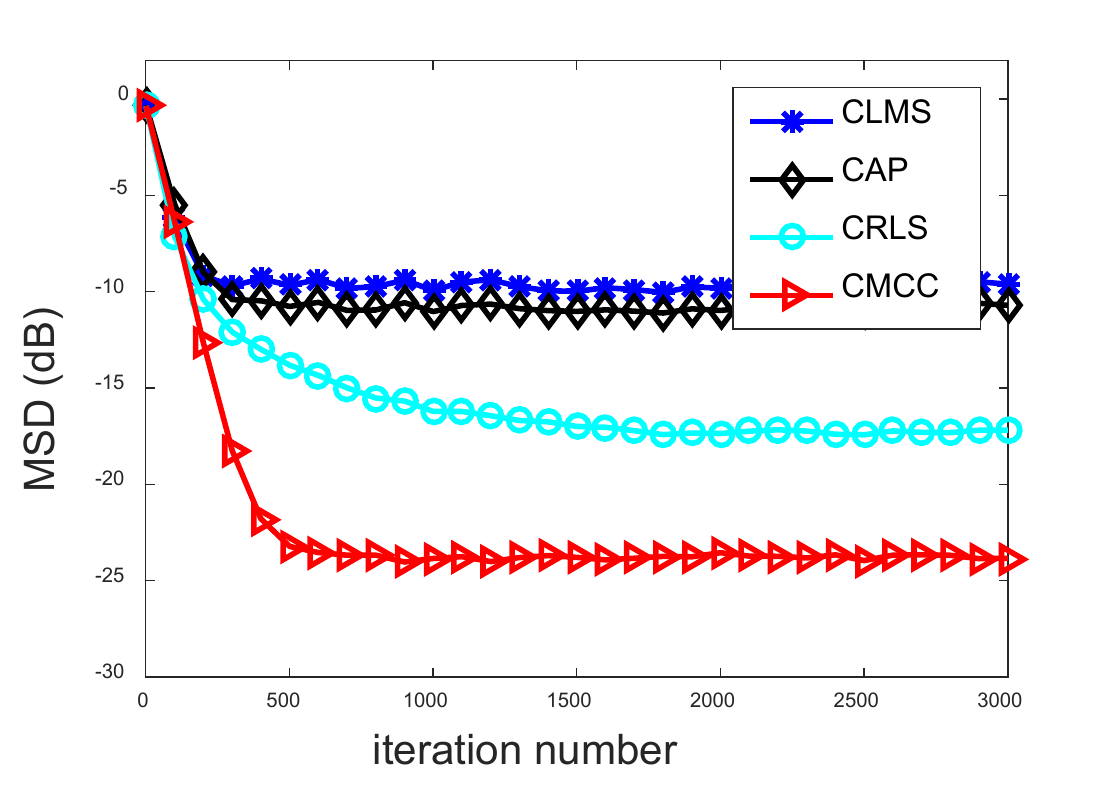}
\label{ls_f1_mixgau}}
\hfil
\subfloat{\includegraphics[width=2.5in]{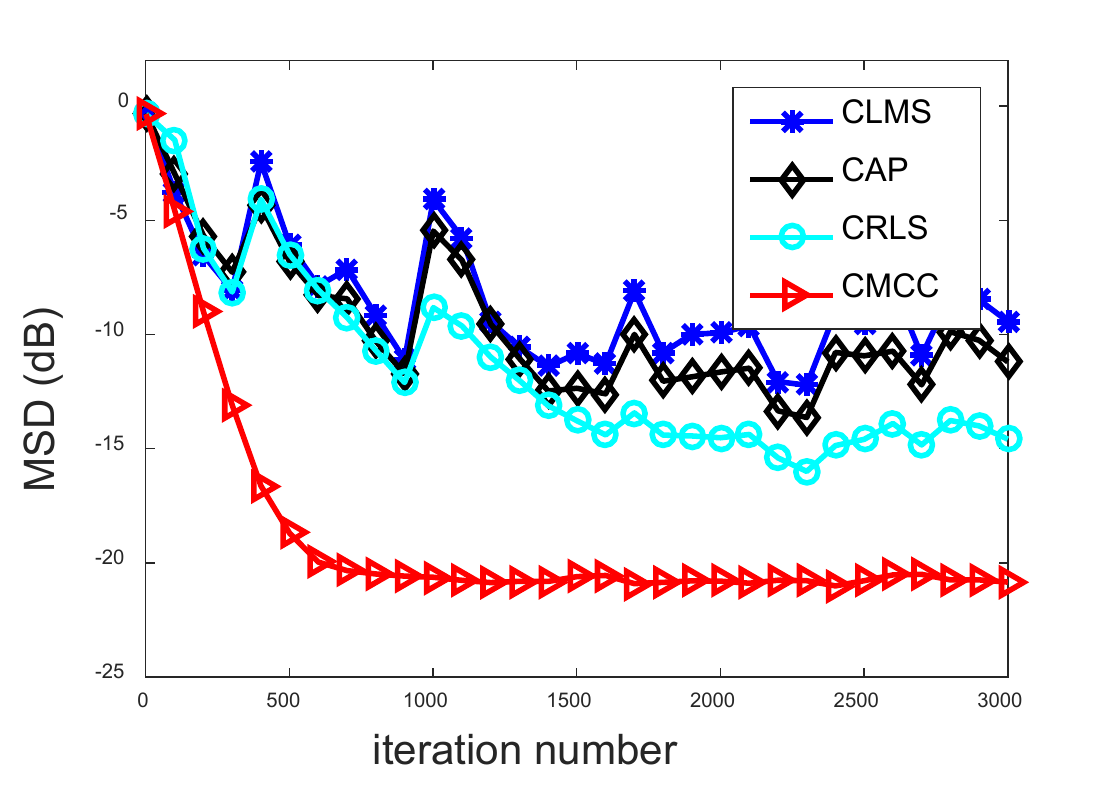}
\label{ls_f1_alpha}}
\hfil
\\(a) \qquad\qquad\qquad\qquad\qquad\qquad\qquad\qquad\qquad\qquad\qquad (b)\\
\subfloat{\includegraphics[width=2.5in]{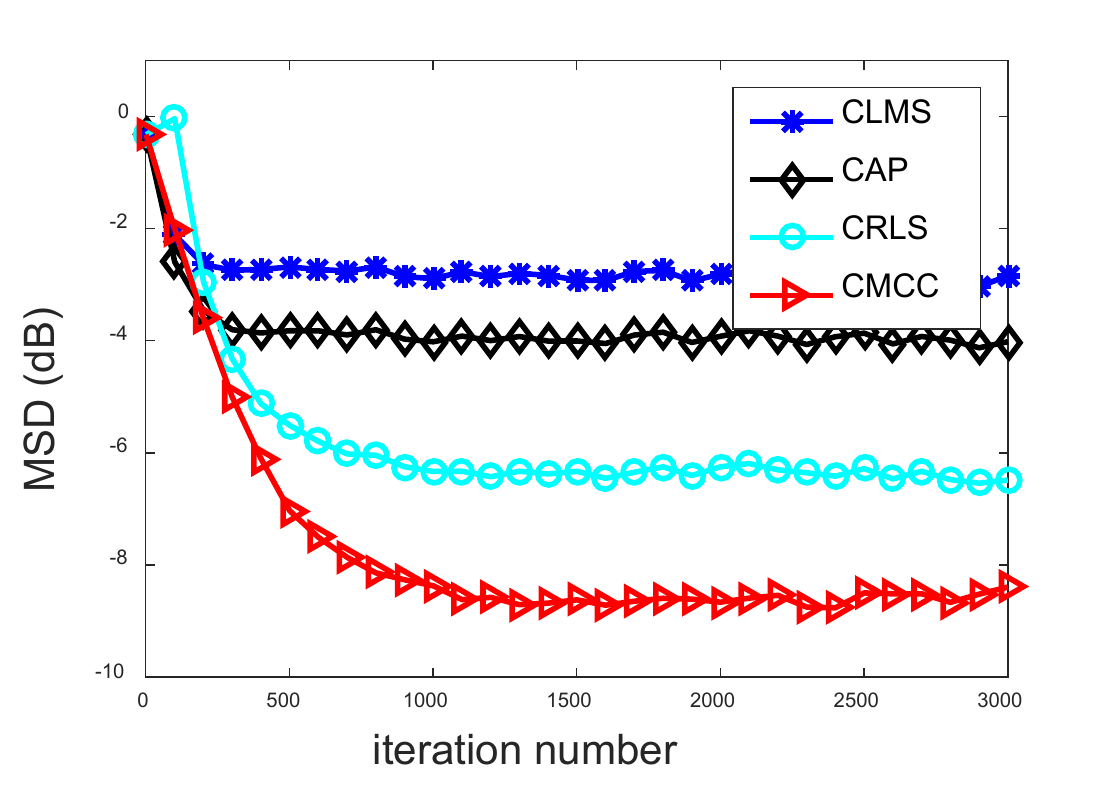}
\label{ls_f1_laplace}}
\hfil
\subfloat{\includegraphics[width=2.5in]{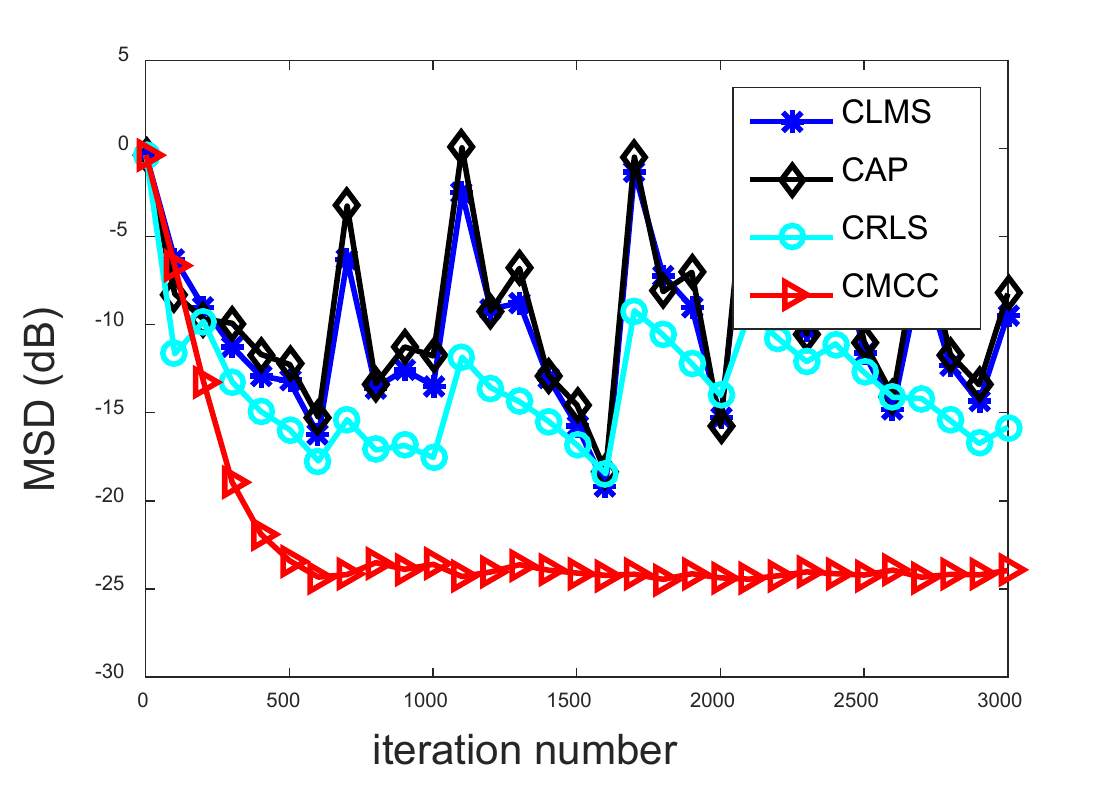}
\label{ls_f1_cauchy}}
\\(c) \qquad\qquad\qquad\qquad\qquad\qquad\qquad\qquad\qquad\qquad\qquad (d)\\
\caption{Convergence curves of CLMS, CAP, CRLS and CMCC in different noises: (a) Mixed Gaussian noise; (b) Alpha-stable noise; (c) Laplace noise; (d) Cauchy noise}
\label{s1_f1}
\end{figure*}
\begin{figure}[!t]
\centering
\includegraphics[width=2.5in]{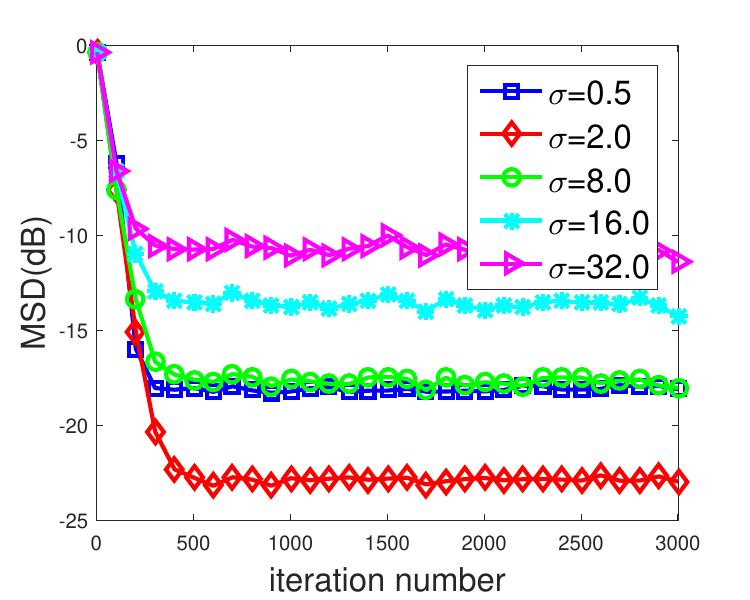}
\caption{Convergence curves of CMCC with different $\sigma$}
\label{ls_f2_kernel}
\end{figure}
\begin{figure}[!t]
\centering
\includegraphics[width=2.5in]{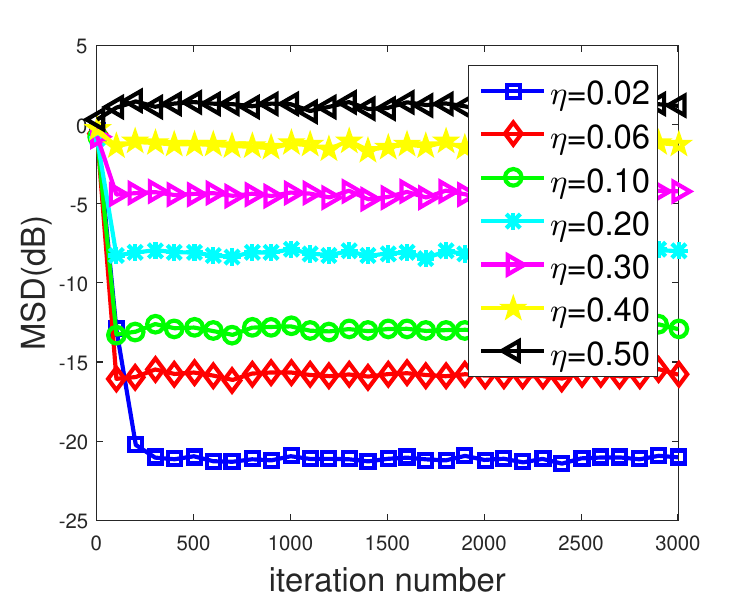}
\caption{Convergence curves of CMCC with different $\eta$}
\label{ls_f3_stepsize}
\end{figure}
\begin{figure}[!t]
\centering
\includegraphics[width=2.5in]{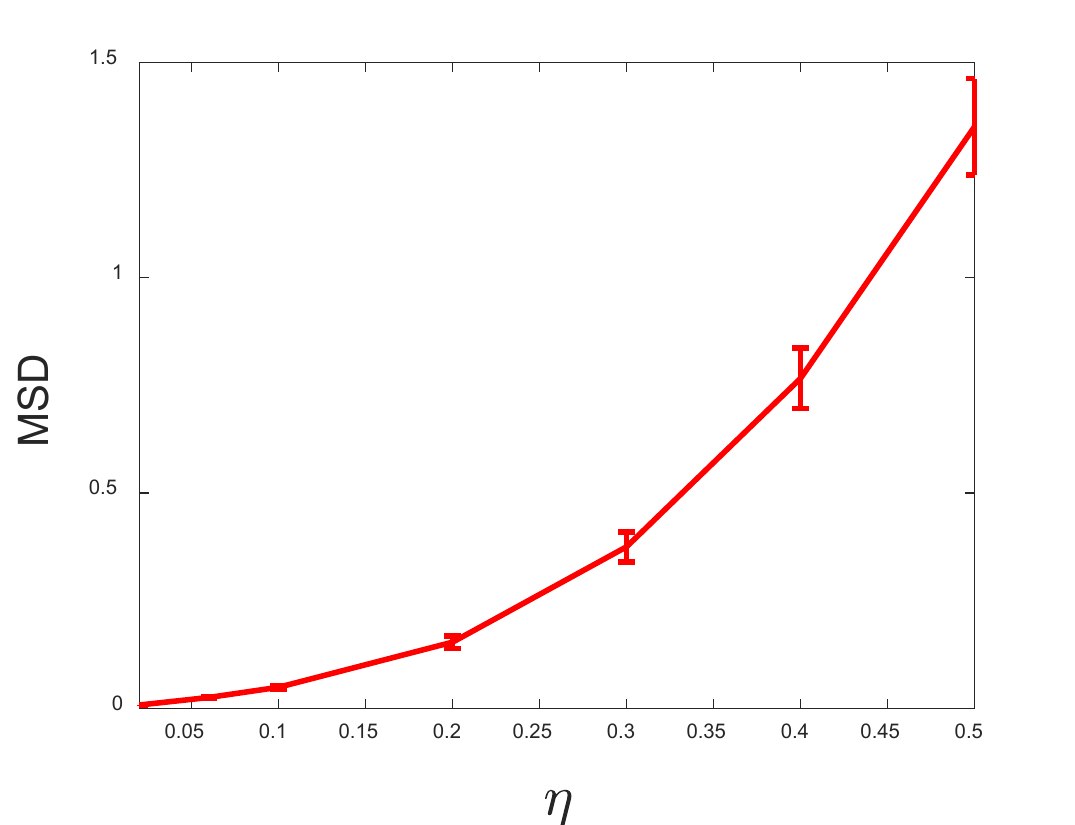}
\caption{Performance evolution curve of CMCC with different $\eta$.}
\label{ls_f3_errorbar}
\end{figure}
\subsection{Linear System Identification}
We consider a linear system identification problem where the length of the adaptive filter is equal to that of the unknown system impulse response. Assume that the weight vector $W^{*}$ of the unknown system, the constraint parameters $C$ and $f$, the input vectors, and the input covariance matrix $\textbf{R}$ are the same as the previous experiment. In the simulations below, without mentioning otherwise, 500 independent Monte Carlo simulations are performed and in each simulation, 3000 iterations are run to ensure the algorithms to reach the steady state. The sliding data length for CAP is set to 4, and the forgetting factor for CRLS is set to 0.998. The kernel bandwidth for CMCC is $\sigma=2.0$. \par
First, we illustrate the performance of the proposed CMCC compared with CLMS, CAP and CRLS in four noise distributions. Simulation results are shown in Fig. \ref{s1_f1}. In the simulation, the mixed Gaussian noise parameters are set at $V_{mix}=\left(0, 0, 0.01, 100, 0.05\right)$, the alpha-stable noise parameters are set as $V_{alpha}=(1.5,0,0.4,0)$, the laplace noise is zero-mean with standard deviation 5, and the cauchy noise is reduced to $\frac{1}{10}$. The step-sizes are chosen such that all the algorithms have almost the same initial convergence speed.  As one can see clearly, the CMCC algorithm significantly outperforms other algorithms in terms of stability, and achieves much lower steady-state MSD. \par
Second, we demonstrate how the kernel bandwidth $\sigma$ will influence the convergence performance of CMCC. Fig. \ref{ls_f2_kernel} shows the convergence curves of CMCC with different $\sigma$, where the mixed gaussian noise is chosen for measurement noise and the noise parameters are the same as the previous simulation. The step-sizes are set at $\eta=0.06, 0.012, 0.01, 0.01, 0.01$ for $\sigma=0.5, 2.0, 8.0, 16.0, 32.0$ respectively. Obviously, the kernel bandwidth has significant influence on the convergence behavior. In this example, the proposed algorithm achieves the lowest steady-state MSD when $\sigma=2.0$. If the kernel bandwidth is too larger (e.g. $\sigma=32.0$) or too small (e.g. $\sigma=0.5$), the convergence performance of CMCC will become poor. We provide some useful properties later for kernel bandwidth selection in practical applications. \par
%
Third, we investigate the stability problem of the CMCC in different step-sizes $\eta$. Fig. \ref{ls_f3_stepsize} illustrates the convergence performance with different step-sizes, and accordingly Fig. \ref{ls_f3_errorbar} shows the performance evolution curve. The noise is still the mixed Gaussian noise with same parameters. From simulation results, one can observe clearly that: 1) when the step-size is very large (such as $\eta\geq0.5$), the CMCC will be divergent, which confirms the validity of the theoretical analysis of mean square stability in section III; 2) As the step-size increases, the mean and variance of MSD of the proposed algorithm become larger. Simulation results show that a larger step-size leads to a more unstable algorithm, and even make the new algorithm to become diverge. Additionally, in this simulation, we calculate the value of $\frac{2}{2\lambda_{max}+tr\left\{\Upsilon\right\}}$ (by (\ref{t_1_15})) to 0.278, not larger than 0.4, which also illustrates the effectiveness of (\ref{t_1_15}).
\subsection{Beamforming Application}
In this scenario, we consider a uniform linear array consisting of $M=7$ omnidirectional sensors with an element spacing of half wavelength. We also assume that there are four users. Among them, the signal of one user is of interest, and is presumed to arrive at the direction-of-arrival (DOA) of $\varphi_{d}=0^{\circ}$, while the other three signals are considered as interferers with DOAs of $\varphi_{1}=-25^{\circ}$, $\varphi_{2}=30^{\circ}$, $\varphi_{3}=60^{\circ}$, respectively. We choose the constraint matrix $C=\left[\textbf{I}_{\frac{M-1}{2}}, \textbf{0}, -\textbf{J}_{\frac{M-1}{2}}\right]$ with $\textbf{J}$ being a reversal matrix of size (an identity matrix with all rows in reversed order), and the response vector $f=\textbf{0}$ \cite{13}. The measurement noise $\upsilon(n)$ is the additive non-Gaussian noise, and the measured output of the unknown system is set to $d(n)=\upsilon(n)$ \cite{11}. In the following simulations, simulation results are averaged over 1000 independent Monte Carlo runs, and in each simulation, 3000 iterations are run to ensure the algorithms to reach the steady state, and the steady-state MSDs are obtained as averages over the last 200 iterations. The \emph{signal-to-noise ratio} (SNR) is set to 0 dB, and the \emph{interference-to-noise ratio} (INR) is set to 10 dB. The sliding data length for CAP is set to 4, and the forgetting factor for CRLS is set to 0.999. The kernel bandwidth $\sigma$ is set at 20.  \par
\begin{figure}[!t]
  \centering
  \includegraphics[width=2.5in]{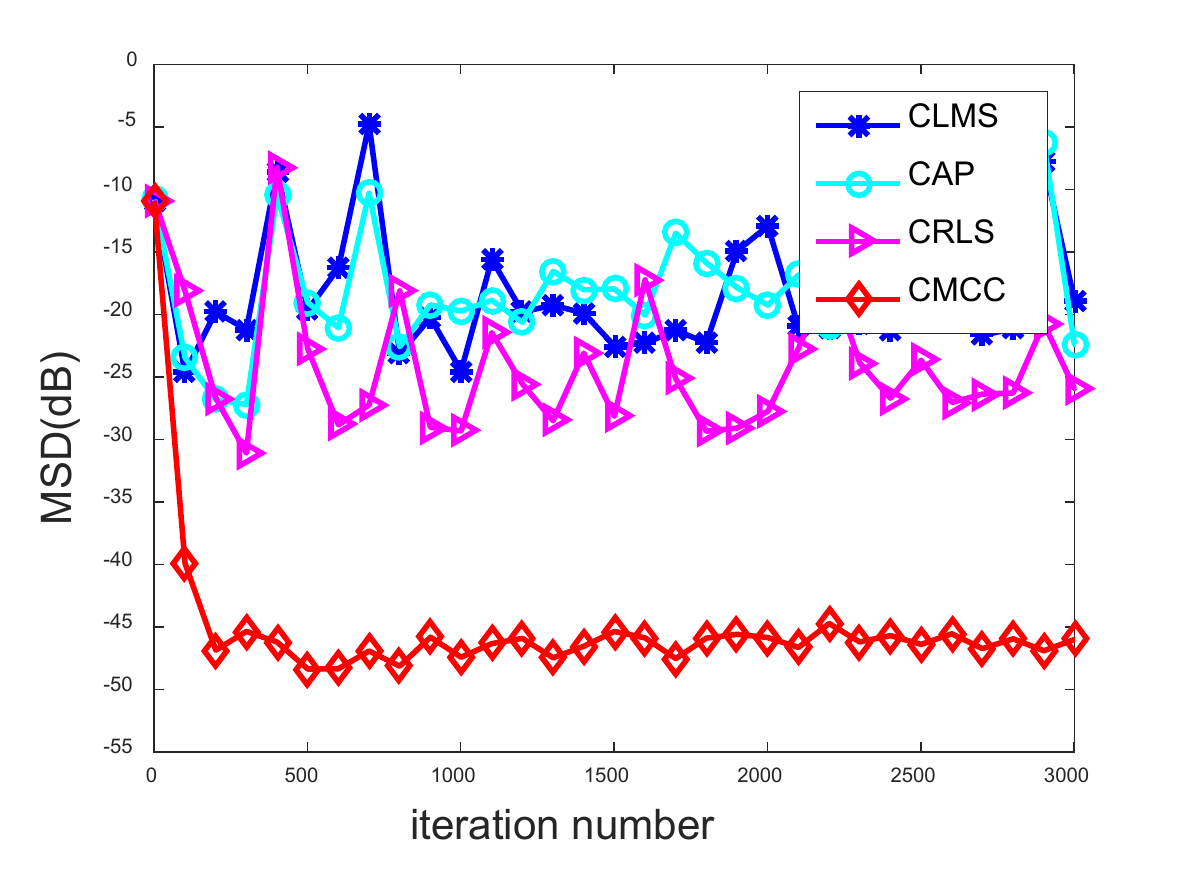}
  \caption{Convergence curves of CLMS, CAP, CRLS and CMCC.}
  \label{be_b1}
\end{figure}
\begin{figure}[!t]
  \centering
  \includegraphics[width=2.5in]{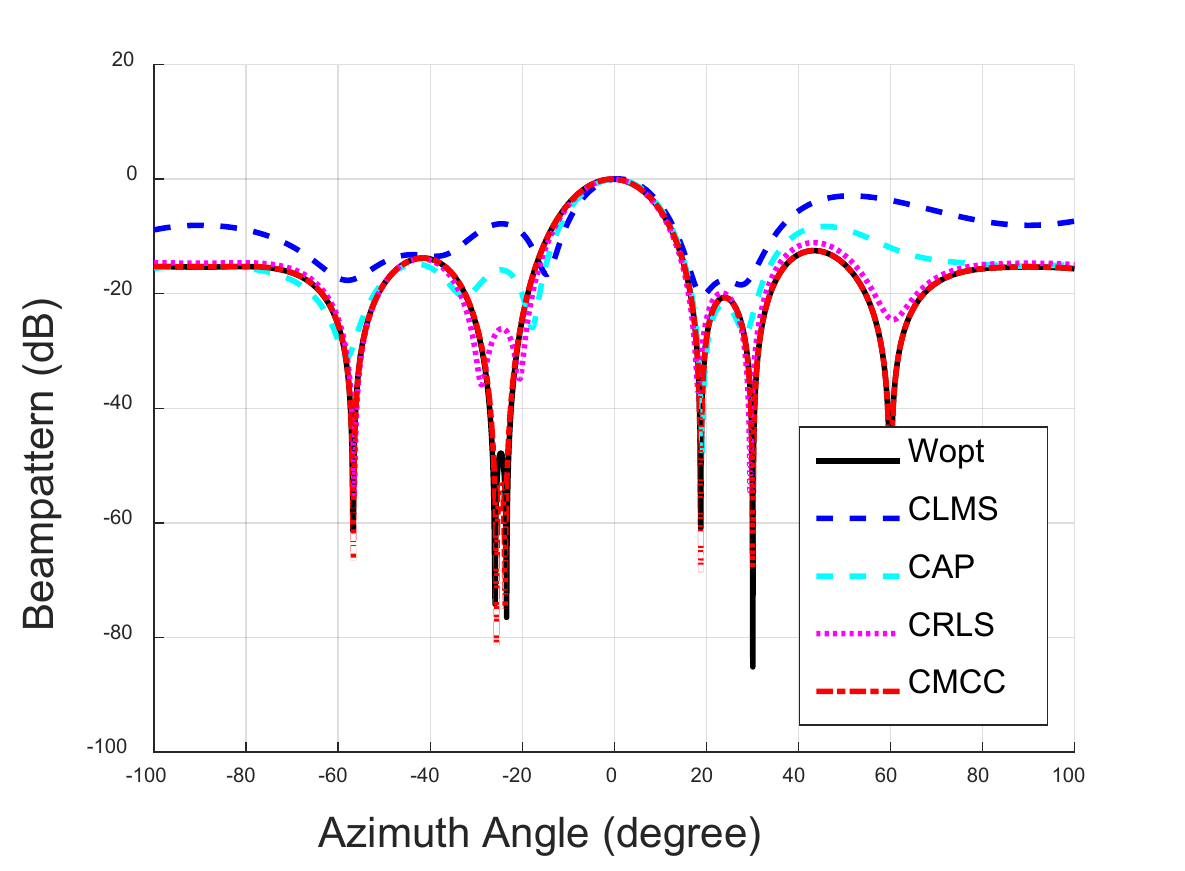}
  \caption{ Beampatterns of CLMS, CAP, CRLS and CMCC.}
  \label{be_b2}
\end{figure}

The convergence curves of CLMS, CAP, CRLS and CMCC in alpha-stable noise are illustrated in Fig. \ref{be_b1}, and accordingly, the beampatterns of different methods are given in Fig. \ref{be_b2}.  The noise parameters are set at $V_{alpha}=(1.2,0,1.6,0)$, and other parameters are chosen such that all algorithms have almost the same initial convergence rate. As one can see that, compared with these traditional constrained adaptive filtering algorithms, the proposed algorithm performs best in all scenarios in term of MSD and beampattern shape. Furthermore, it has similar performance to the optimal beamformer after convergence.\par
\begin{figure}[!t]
  \centering
  \includegraphics[width=3.0in]{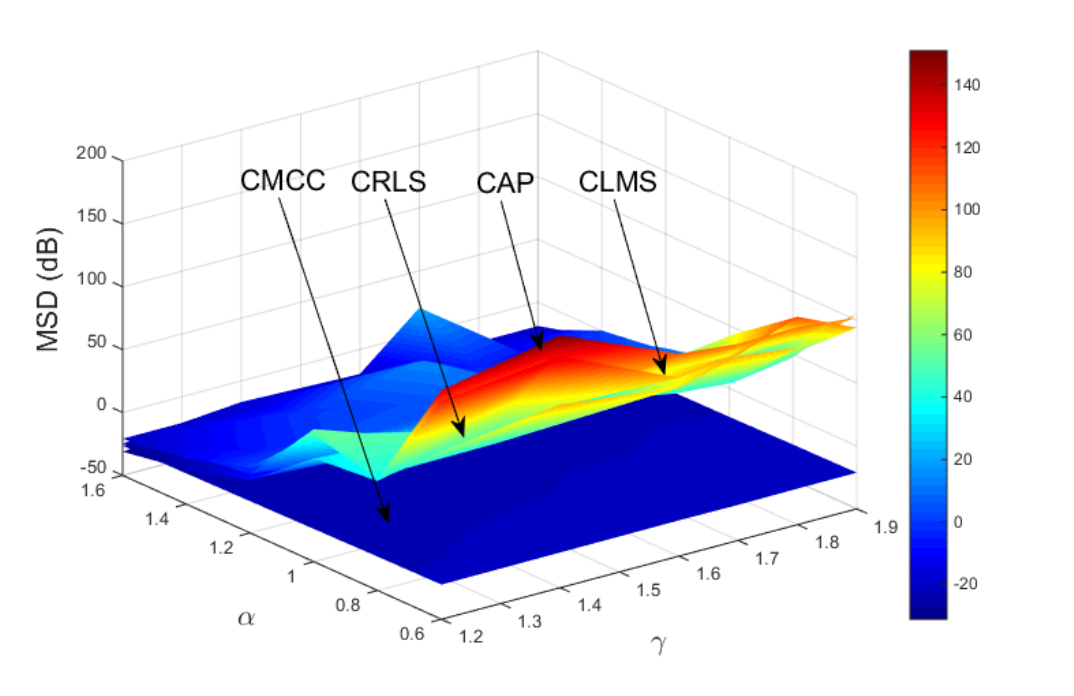}
  \caption{Steady-state MSDs of CLMS, CAP, CRLS and CMCC in 3-D space.}\label{fig9}
  \label{s2_3}
\end{figure}

Fig. \ref{s2_3} shows the steady-state MSDs of CLMS, CAP, CRLS and CMCC with different $\alpha=(0.6, 0.8, 1.0, 1.2, 1.4, 1.6)$ and different $\gamma=(1.2, 1.4, 1.6, 1.7, 1.8, 1.9)$ in 3-D space. Other parameters are the same as in the previous simulation for all algorithms. As expected, the proposed algorithm can achieve much better steady-state performance than CLMS, CAP and CRLS in all cases.

\subsection{Parameter Selection}
The kernel bandwidth $\sigma$ is an important free parameter in CMCC since it controls all robust properties of correntropy. An appropriate kernel bandwidth can provide an effective mechanism to eliminate the effect of outliers and noise.

According to the previous studies, some useful tricks for kernel bandwidth selection are as follows \cite{23, 24, 25, 26, 27, 28, 29, 30}:
\begin{enumerate}
  \item If the data are plentiful, a small $\sigma$ should be used so that high precision can be achieved; however, the kernel bandwidth must be selected to make a compromise between estimation efficiency and outlier rejection if the data are small.
  \item As $\sigma$ increases, the contribution of the higher-order moments decays faster, and the second-order moment plays a key role. Therefore, a large $\sigma$ is frequently appropriate for Gaussian noises, while a small $\sigma$ is usually adapt to non-Gaussian impulsive noises.
  \item For a given noise environment, there is a relatively large range of $\sigma$ that provides nearly optimal performance.
\end{enumerate} \par
Currently, Silverman's rule, one of the most widely used methods in kernel density estimation, is often used to estimate $\sigma$. However, the limitation is that this method cannot obtain the best possible value. Therefore, in a practical application, $\sigma$ is manually selected or optimized by trials and errors.
%
%
\section{Conclusion}
In this paper, we have developed the constrained maximum correntropy criterion (CMCC) adaptive filtering algorithm by incorporating a linear constraint into the maximum correntropy criterion. We also studied the mean square convergence performance including the mean square stability and the steady-state mean square deviation (MSD) of the proposed algorithm. Simulation results have confirmed the theoretical conclusions and shown that the new algorithm can significantly outperform the traditional CLMS, CAP and CRLS algorithms when the noise is of heavy-tailed non-Gaussian distribution. \par
A main benefit of correntropy is that the kernel bandwidth controls all its properties. However, in practical applications, the kernel bandwidth is manually selected by scanning the performance. Therefore, how to select an optimal kernel bandwidth is a big challenge for future study. On the other hand, CMCC belongs to the family of stochastic gradient based algorithms, which usually suffers from slow convergence. It is expected to solve this problem by investigating the MCC-based constrained affine projection algorithm and recursive maximum correntropy algorithm.

%
%
\appendices
\section{Derivation of (\ref{c_2_7})}
Based on (\ref{c_2_6_2}), we can easily derive the following instantaneous weight update equation \cite{24, 29}
\begin{align}\label{app1-2}
W(n)&=W(n-1)+\eta\frac{\partial J_{CMCC}}{\partial W}|_{W=W(n-1)} \nonumber \\
    &=W(n-1)+\eta g(e(n))\left(d(n)-W^{T}(n-1)X(n)\right)\nonumber \\
    &\;\;\;\;\times X(n)+\eta C\xi(n)
\end{align}
where the weight vector $W$ is initialized at a vector satisfying $W(0)=C(C^{T}C)^{-1}f$ . Due to
\begin{align}\label{app1-3}
f=&C^{T}W(n) \nonumber \\
 =&C^{T}\left[W(n-1)+\eta C\xi(n)+\eta g(e(n))\times\right. \nonumber \\
   &\left.\left(d(n)-W^{T}(n-1)X(n)\right)X(n)\right]
\end{align}
we have
\begin{align}\label{app1-4}
\xi(n)=&\frac{1}{\eta}(C^{T}C)^{-1}
        \left[f-C^{T}W(n-1)-\eta g(e(n))\times\right. \nonumber\\
        &\left.\left(d(n)-W^{T}(n-1)X(n)\right)C^{T}X(n)\right]
\end{align}
Substituting (\ref{app1-4}) into (\ref{app1-2}), and after some simple vector manipulations, we derive
\begin{align}\label{app1-5}
W(n)=&P\left[W(n-1)+\eta g(e(n))(d(n)-\right. \nonumber \\
     & \left.W^{T}(n-1)X(n))X(n)\right]+Q
\end{align}
which is the CMCC algorithm.

\section{Derivation of (\ref{t_1_2})}
Setting $\frac{\partial J_{CMCC}}{\partial W}|_{W=W(n-1)}=\textbf{0}_{M\times1}$, one can derive the optimal weight vector $W_{opt}$ under CMCC as follows:
\begin{align}\label{app2-1}
&E\left[g(e(n))(d(n)-W^{T}_{opt}X(n))X(n)\right]+C\xi(n)=\textbf{0}_{M\times1}  \nonumber \\
&\Rightarrow E\left[g(e(n))X(n)X^{T}(n)\right]W_{opt}=E\left[g(e(n))d(n)X(n)\right]+  \nonumber \\
&\qquad\qquad\qquad\qquad\qquad\qquad\qquad\qquad C\xi(n)                                             \nonumber \\
&\Rightarrow \textbf{R}_{g}W_{opt}=\textbf{P}_{g}+C\xi(n)   \nonumber \\
&\Rightarrow  W_{opt}= \textbf{R}_{g}^{-1}\textbf{P}_{g}+\textbf{R}_{g}^{-1}C\xi(n)
\end{align}
where $\textbf{P}_{g}=E\left[g(e(n))d(n)X(n)\right]$ is a weighted cross-correlation vector between the measured output and the input vector.
Since
\begin{align}\label{app2-2}
&C^{T}W_{opt}=f  \nonumber \\
&\Rightarrow C^{T}\left[\textbf{R}_{g}^{-1}\textbf{P}_{g}+\textbf{R}_{g}^{-1}C\xi(n)\right]=f \nonumber \\
&\Rightarrow \xi(n)=\left[C^{T}\textbf{R}_{g}^{-1}C\right]^{-1}\left(f-C^{T}\textbf{R}_{g}^{-1}\textbf{P}_{g}\right)
\end{align}
one can rewrite (\ref{app2-1}) as
\begin{align}\label{app2-3}
W_{opt}=&\textbf{R}_{g}^{-1}\textbf{P}_{g}+\textbf{R}_{g}^{-1}C\left[C^{T}\textbf{R}_{g}^{-1}C\right]^{-1}\times  \nonumber \\
        &\left(f-C^{T}\textbf{R}_{g}^{-1}\textbf{P}_{g}\right)
\end{align}
Under the assumptions 1) and 2), we derive by using (\ref{c_2_1})
\begin{align}\label{app2-4}
&d(n)=W^{*T}X(n)+\upsilon(n)         \nonumber \\
&\Rightarrow  d(n)X^{T}(n)=W^{*T}X(n)X^{T}(n)+\upsilon(n)X^{T}(n) \nonumber \\
&\Rightarrow  g(e(n))d(n)X^{T}(n)=g(e(n))W^{*T}X(n)X^{T}(n)+  \nonumber \\
& \qquad\qquad \qquad \qquad \qquad \quad               \upsilon(n)g(e(n))X^{T}(n) \nonumber \\
&\Rightarrow  \textbf{P}_{g}=\textbf{R}_{g}W^{*}   \nonumber \\
&\Rightarrow  W^{*}=\textbf{R}_{g}^{-1}\textbf{P}_{g}
\end{align}
Therefore, combining (\ref{app2-3}) and (\ref{app2-4}), we obtain
\begin{align}\label{app2-5}
{W}_{opt}=W^{*}+\textbf{R}_{g}^{-1}C\left(C^{T}\textbf{R}_{g}^{-1}C\right)^{-1}\left(f-C^{T}W^{*}\right)
\end{align}
%

\section{Derivation of (\ref{t_2_13})$\sim$(\ref{t_2_18})}

Here we consider two cases below:
\begin{enumerate}
  \item Gaussian noise case \\
        Since $e(n)=e_{a}(n)+\upsilon(n)$, in this case $e(n)$ is also zero-mean Gaussian. Let $\sigma^{2}_{e}$ be the variance of the error $e(n)$. Then we have
        \begin{align}\label{app3-1}
        \sigma^{2}_{e}=E\left[e^{2}_{a}(n)\right]+\sigma^{2}_{\upsilon}
        \end{align}
        Using (\ref{t_1_3}) and the approximation $W(n)\approx W_{opt}$ at the steady-state, we obtain
        \begin{align}\label{app3-2}
        e_{a}(n)\approx\left(W^{\ast}-W_{opt}\right)^{T}X(n)=\varepsilon_{w} X(n)
        \end{align}
        Therefore
        \begin{align}\label{app3-3}
        \sigma^{2}_{e}\approx\varepsilon^{T}_{w}\textbf{R}\varepsilon_{w}+\sigma^{2}_{\upsilon}
        \end{align}
        It follows that
        \begin{align}\label{app3-4}
        \lim\limits_{n\rightarrow\infty}E\left[g(e(n))\right]&=\lim\limits_{n\rightarrow\infty}\frac{1}{\sqrt{2\pi}\sigma_{e}}\int\limits_{-\infty}^{\infty}
        \exp\left(-\frac{e^{2}(n)}{2\sigma^{2}}\right)\times\nonumber \\
        &\;\;\;\;\;\;\exp\left(-\frac{e^{2}(n)}{2\sigma_{e}^{2}}\right)de(n) \nonumber \\
        &=\frac{\sigma}{\sqrt{\sigma^{2}+\sigma^{2}_{e}}}   \nonumber \\
        &\approx\frac{\sigma}{\sqrt{\sigma^{2}+\varepsilon^{T}_{w}\textbf{R}\varepsilon_{w}+\sigma_{\upsilon}^{2}}}
        \end{align}
        %
        \begin{align}\label{app3-5}
        \lim\limits_{n\rightarrow\infty}E\left[g^{2}(e(n))\right]&=\lim\limits_{n\rightarrow\infty}\frac{1}{\sqrt{2\pi}\sigma_{e}}\int\limits_{-\infty}^{\infty}
        \exp\left(-\frac{e^{2}(n)}{\sigma^{2}}\right)\times\nonumber \\
        &\;\;\;\;\;\;\exp\left(-\frac{e^{2}(n)}{2\sigma_{e}^{2}}\right)de(n) \nonumber \\
        &=\frac{\sigma}{\sqrt{\sigma^{2}+2\sigma^{2}_{e}}}   \nonumber \\
        &\approx\frac{\sigma}{\sqrt{\sigma^{2}+2\varepsilon^{T}_{w}\textbf{R}\varepsilon_{w}+2\sigma_{\upsilon}^{2}}}
        \end{align}
        Substituting (\ref{app3-4}) and (\ref{app3-5}) into (\ref{t_2_10}), we obtain
        \begin{align}\label{app3-6}
        S\approx &         \eta^{2}\left(\varepsilon^{T}_{w}\textbf{R}\varepsilon_{w}+\sigma_{\upsilon}^{2}\right)\textbf{vec}^{T}\left\{\Upsilon\right\}
        \left(\textbf{I}_{M^{2}}-\textbf{F}\right)^{-1}\times \nonumber \\
        &\qquad\textbf{vec}\{\textbf{I}_{M}\}\frac{\sigma}{\sqrt{\sigma^{2}+2\varepsilon^{T}_{w}\textbf{R}\varepsilon_{w}+2\sigma_{\upsilon}^{2}}}
        \end{align}
    \item Non-Gaussian noise case \\
        Taking the Taylor expansion of $g(e(n))$ with respect to $e_{a}(n)$ around $\upsilon(n)$, we have
        \begin{align}\label{app3-7}
            g(e(n))=&g(e_{a}(n)+\upsilon(n))   \nonumber \\
                   =&g(\upsilon(n))+g'(\upsilon(n))e_{a}(n)+ \nonumber \\
                   &\frac{1}{2}g''(\upsilon(n))e^{2}_{a}(n)+o(e^{2}_{a}(n))
        \end{align}
        where
        \begin{align}\label{app3-8}
            g(\upsilon(n))=\exp\left(-\frac{\upsilon^{2}(n)}{2\sigma^{2}}\right)
        \end{align}
        %
        \begin{align}\label{app3-9}
            g'(\upsilon(n))=-\frac{\upsilon(n)}{\sigma^{2}}\exp\left(-\frac{\upsilon^{2}(n)}{2\sigma^{2}}\right)
        \end{align}
        %
        \begin{align}\label{app3-10}
            g''(\upsilon(n))=(\frac{\upsilon^{2}(n)}{\sigma^{4}}-\frac{1}{\sigma^{2}})\exp\left(-\frac{\upsilon^{2}(n)}{2\sigma^{2}}\right)
        \end{align}
        Thus
        \begin{align}\label{app3-11}
            E\left[g(e(n))\right]\approx& E\left[g(\upsilon(n))\right]+\frac{1}{2}E\left[g''(\upsilon(n))\right]E\left[e^{2}_{a}(n)\right] \nonumber \\
            =&E\left[\exp\left(-\frac{\upsilon^{2}(n)}{2\sigma^{2}}\right)\right]+\frac{1}{2}\varepsilon^{T}_{w}\textbf{R}\varepsilon_{w}\times      \nonumber \\
             &E\left[\left(\frac{\upsilon^{2}(n)}{\sigma^{4}}-\frac{1}{\sigma^{2}}\right)\exp\left(-\frac{\upsilon^{2}(n)}{2\sigma^{2}}\right)\right]
        \end{align}
        %
        \begin{align}\label{app3-12}
            E\left[g^{2}(e(n))\right]\approx &E\left[g(\upsilon^{2}(n))\right]+E\left[e^{2}_{a}(n)\right]\times \nonumber \\
            &E\left[g(\upsilon(n))g''(\upsilon(n))+g'^{2}(\upsilon(n)))\right] \nonumber \\
            =&E\left[\exp\left(-\frac{\upsilon^{2}(n)}{\sigma^{2}}\right)\right]+\varepsilon^{T}_{w}\textbf{R}\varepsilon_{w}\times \nonumber \\
            &E\left[\left(\frac{2\upsilon^{2}(n)}{\sigma^{4}}-\frac{1}{\sigma^{2}}\right)\exp\left(-\frac{\upsilon^{2}(n)}{\sigma^{2}}\right)\right]
        \end{align}
        Substituting (\ref{app3-11}) and (\ref{app3-12}) into (\ref{t_2_10}) yields
        \begin{align}\label{app3-13}
                S\approx& \eta^{2}\left(\varepsilon^{T}_{w}\textbf{R}\varepsilon_{w}+E\left[\upsilon^{2}(n)\right]\right)\textbf{vec}^{T}\left\{\Upsilon\right\} \left(\textbf{I}_{M^{2}}-\textbf{F}\right)^{-1}\times \nonumber\\
                &\textbf{vec}\{\textbf{I}_{M}\}\left(E\left[\exp\left(-\frac{\upsilon^{2}(n)}{\sigma^{2}}\right)\right]+
                \varepsilon^{T}_{w}\textbf{R}\varepsilon_{w}\times \right.\nonumber \\
                &\left.E\left[(\frac{2\upsilon^{2}(n)}{\sigma^{4}}-\frac{1}{\sigma^{2}}\right)\exp\left(-\frac{\upsilon^{2}(n)}{\sigma^{2}}\right)\right])
        \end{align}
\end{enumerate}

\section*{Acknowledgments}

This work was supported by 973 Program (No. 2015CB351703) and National Natural Science Foundation of China (No. 61372152).

\ifCLASSOPTIONcaptionsoff
  \newpage
\fi

\end{document}